\documentclass[10pt,twocolumn,letterpaper]{article}

\usepackage[pagenumbers]{cvpr} % To force page numbers, e.g. for an arXiv version

\definecolor{cvprblue}{rgb}{0.21,0.49,0.74}
\usepackage[pagebackref,breaklinks,colorlinks,allcolors=cvprblue]{hyperref}
\usepackage{multirow}
\usepackage{bbding}
\usepackage{algorithm}
\usepackage{listings}
\usepackage{xcolor}
\usepackage{float}
\usepackage{colortbl}
\definecolor{gray}{rgb}{0.8,0.8,0.8}
\usepackage{etoolbox}
\makeatletter
\AfterEndEnvironment{algorithm}{\let\@algcomment\relax}
\AtEndEnvironment{algorithm}{\kern2pt\hrule\relax\vskip3pt\@algcomment}
\let\@algcomment\relax
\newcommand\algcomment[1]{\def\@algcomment{\footnotesize#1}}
\renewcommand\fs@ruled{\def\@fs@cfont{\bfseries}\let\@fs@capt\floatc@ruled
  \def\@fs@pre{\hrule height.8pt depth0pt \kern2pt}%
  \def\@fs@post{}%
  \def\@fs@mid{\kern2pt\hrule\kern2pt}%
  \let\@fs@iftopcapt\iftrue}
\makeatother
\def\paperID{****} % *** Enter the Paper ID here
\def\confName{CVPR}
\def\confYear{2026}

\title{Repulsor: Accelerating Generative Modeling with a Contrastive Memory Bank}

\author{Shaofeng Zhang$^{1*\dagger}$, Xuanqi Chen$^{1}$\thanks{: Equal contribution. $\dagger$: Corresponding author}, Ning Liao$^{2*}$, Haoxiang Zhao$^1$ \\
Xiaoxing Wang$^2$, Haoru Tan$^3$, Sitong Wu$^4$, Xiaosong Jia$^5$, Qi Fan$^6$, Junchi Yan$^2$ \\
sfzhang@ustc.edu.cn, yanjunchi@sjtu.edu.cn \\
$^1$School of Artificial Intelligence and Data Science, University of Science and Technology of China, \\
$^2$ Shanghai Jiao Tong University, $^3$HKU, $^4$CUHK, $^5$Fudan University, $^6$Nanjing University
}

\newcommand{\mname}{Repulsor}

\begin{document}
% \maketitle
% \begin{figure*}[ht]
%     \centering
%     \includegraphics[width=.96\linewidth]{figures/fig1.pdf}
%     \caption{\textbf{Selected samples on ImageNet 256$\times$256} from the SiT-XL/2 + {\mname} model. We use classifier-free guidance with $w=4.0$. 
%     }
%     \label{fig:main_qual}
%     \vspace{-0.15in}
% \end{figure*}

\maketitle

% \twocolumn[{
% \renewcommand\twocolumn[1][]{#1}
% \maketitle
% \begin{center}
%     \vspace{-16pt}
%     \includegraphics[width=1.0\linewidth]{figures/fig1.pdf}
%     \vspace{-20pt}
%     \captionsetup{type=figure}
%     \caption{Selected samples from the SiT-XL/2+{\mname} on ImageNet 256×256, generated at 400k training iterations.
%     }
%     \label{fig:teaser}
%     \vspace{-2pt}
% \end{center}
% }]

\begin{abstract}
    The dominance of denoising generative models (e.g., diffusion, flow-matching) in visual synthesis is tempered by their substantial training costs and inefficiencies in representation learning. While injecting discriminative representations via auxiliary alignment has proven effective, this approach still faces key limitations: the reliance on external, pre-trained encoders introduces overhead and domain shift. A dispersed-based strategy that encourages strong separation among in-batch latent representations alleviates this specific dependency. To assess the effect of the number of negative samples in generative modeling, we propose {\mname}, a plug-and-play training framework that requires no external encoders. Our method integrates a memory bank mechanism that maintains a large, dynamically updated queue of negative samples across training iterations. This decouples the number of negatives from the mini-batch size, providing abundant and high-quality negatives for a contrastive objective without a multiplicative increase in computational cost. A low-dimensional projection head is used to further minimize memory and bandwidth overhead. {\mname} offers three principal advantages: (1) it is self-contained, eliminating dependency on pretrained vision foundation models and their associated forward-pass overhead; (2) it introduces no additional parameters or computational cost during inference; and (3) it enables substantially faster convergence, achieving superior generative quality more efficiently. On ImageNet-256, {\mname} achieves a state-of-the-art FID of \textbf{2.40} within 400k steps, significantly outperforming comparable methods.
    
\end{abstract}

\section{Introduction}

% Recent Progress of Diffusion models

% Self-supervised contrastive learning methods

% Connection of contrastive model and diffusion models, like RePA, disperse and its key limitation

% Inspired by moco, introduce memory bank. Our method enjoys three advantages: i) {\mname} does not require pretrained visual foundation models, \eg DINO v2. ii) {\mname} does not introduce any parameters and extra computational cost in inference stage, and iii) {\mname} show much faster convergence speed than previous methods~\cite{disperse, repa}. Specifically, our {\mname} achieves xxx FID, surpassing xxx.

In recent years, generative models operating on denoising principles, such as diffusion \cite{ldm, ddpm, ddim, sde, beats} and flow-based \cite{cfm, rectflow, voicebox, boosting_flow_matching} models, have become the predominant and scalable paradigm for high-dimensional visual modeling. They have demonstrated exceptional performance in core applications, including video generation \cite{dit, sora, sd3}, text-to-image synthesis \cite{ldm, dalle2, imagen}, and image generation \cite{controlnet, animate_anyone}. However, substantial training costs and bottlenecks in representation learning efficiency continue to impede further progress. A promising and validated approach to address these limitations involves incorporating an auxiliary representation learning task. This method aligns the intermediate features of the generator with a semantically rich and structurally robust representation space, which serves to regularize the feature manifold and ease the optimization process, especially during the early training.

Self-supervised contrastive learning~\cite{zha2023rank, bao2021beit,chen2024context, he2020momentum,grill2020bootstrap,caron2020unsupervised,chen2021exploring,zhang2022dino} leverages data augmentation to construct positive pairs from augmented views of the same instance, while views from different instances serve as negative samples. The learning process is driven by optimizing an InfoNCE-based objective~\cite{tian2020contrastive}, which encourages the model to attract positive pairs in the embedding space while repelling negative ones. This paradigm results in representations that are not only semantically discriminative and linearly separable but also demonstrate fast convergence speed~\cite{he2022masked, xie2022simmim}.

\begin{figure}
    \centering
    \includegraphics[width=\linewidth]{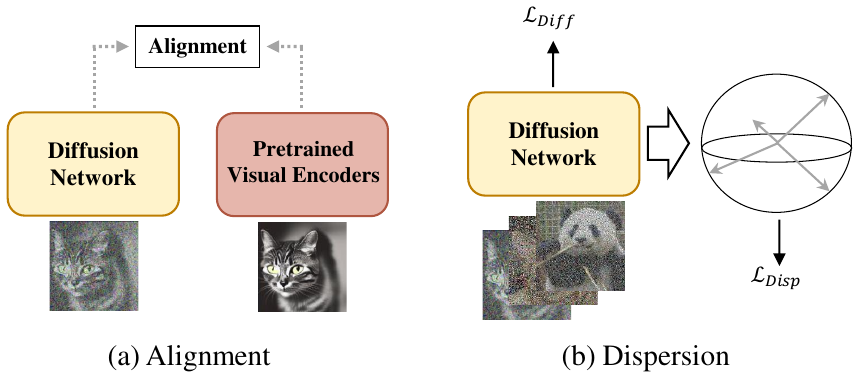}
    \caption{Two primary strategies for incorporating discriminative representations into the training of diffusion models: (a) Representation alignment guided notably by an external pre-trained encoder, exemplified by REPA \cite{repa}; and (b) Positive-free contrastive regularization, as previously introduced in disperse \cite{Disperse}.}
    \label{fig:align_w_dispersion}
\end{figure}

To speed up the training of diffusion models, recent efforts to inject discriminative representations into diffusion training have progressed along two directions (see Figure~\ref{fig:align_w_dispersion}): alignment \cite{repa,SARA,U-REPA,REPA-E,SoftREPA,VA-VAE,xu2025exploring} and dispersion \cite{Disperse}. REPA \cite{repa} introduces an auxiliary alignment loss without modifying the primary objective, mapping noisy intermediate features into the clean semantic space of a large self-supervised encoder, which stabilizes early optimization and accelerates convergence. Although REPA substantially improves training efficiency and generative quality, important limitations persist: REPA relies on an external, pre-trained encoder, which increases forward-pass overhead and introduces the risk of domain shift.
To address this limitation, inspired by negative pair-based contrastive learning, another line based on disperse \cite{Disperse} objective is proposed, which adopts contrastive regularization without positives, encouraging strong separation among in-batch latent representations in feature space to prevent collapse, while remaining jointly optimizable with the diffusion objective. 
% Although both approaches patch-level similarity alignment in REPA provides limited explicit control over structural relations and global distributional consistency. 

A substantial body of work has established that the number of negative samples is pivotal in contrastive learning: more negative pairs strengthen InfoNCE and yield stronger representations ~\cite{he2020momentum, koohpayegani2021mean}. Building on negative-based dispersive loss for generative modeling, we ask: \textbf{Does an analogous scaling law hold for generative models with respect to the number of negative pairs?} However, simply scaling the mini-batch to harvest more negatives conflicts with our goal of efficient and accessible training and often incurs multiplicative cost~\cite{chen2020simple}. We therefore propose {\mname}, which decouples the effective number of negatives from the current mini-batch by maintaining a large, dynamically updated queue-based memory bank across iterations. A lightweight projector is further introduced to map features into a low-dimensional space, reducing memory and bandwidth while preserving discriminative capacity. Our {\mname} has three main advantages:

\textbf{i) {\mname} does not rely on any pretrained vision foundation model (\eg DINO v2 \cite{Oquab2023DINOv2LR}). }By leveraging a memory bank together with a low-dimensional projection head, it constructs abundant and high-quality negatives, learns discriminative representations directly within the generator, and avoids the forward pass overhead and domain shift risks introduced by external encoders.

\textbf{ii) {\mname} introduces no additional parameters or computation at inference. }The memory bank and the projection head are used only during training as auxiliary objectives, and the inference path is identical to the baseline.

\textbf{iii) {\mname} converges substantially faster than existing methods ~\cite{Disperse,sit}. }Under the same training configuration, it reaches superior generative quality earlier. Concretely, on ImageNet-256 it achieves \textbf{2.40 FID} within 400k steps (w/ CFG), surpassing disperse at 5.09 and SiT at 6.02.

\section{Related Work}

\noindent \textbf{Self-Supervised Learning~}~ Self-supervised learning (SSL) in computer vision aims to exploit large collections of unlabeled images by designing proxy signals for pretraining, enabling models to acquire transferable and discriminative representations without manual annotations. Contemporary SSL methods largely fall into two families: contrastive learning (CL) \cite{he2020momentum,grill2020bootstrap,caron2020unsupervised,chen2021exploring} and masked image modeling (MIM) \cite{bao2021beit,chen2024context}. CL methods optimize instance-level invariance by maximizing agreement between embeddings of different augmentations of the same image while pushing apart others, thereby uncovering structure in unlabeled data. Representative approaches include SimCLR \cite{chen2020simple}, which emphasizes strong data augmentation with a simple objective, DINO \cite{zhang2022dino}, which adapts self-distillation to the unlabeled regime, and Rank-N-Contrast (RNC) \cite{zha2023rank}, which introduces a ranking-based objective grounded in sequential representation learning. In contrast, MIM corrupts images by masking patches and trains models to reconstruct the missing content from visible context, encouraging the learning of semantically rich features. MAE \cite{he2022masked} is a canonical example, employing a lightweight autoencoder to reconstruct masked regions. Building on this paradigm, subsequent work proposes new pretraining designs \cite{amac2022masksplit,xie2022simmim} and explores alternative reconstruction targets and objectives \cite{wei2022masked,baevski2022data2vec,dong2023peco}.

\noindent \textbf{Diffusion Models~}~ Recent generative diffusion methods are predominantly categorized into two classes: Diffusion Probabilistic Models (DPMs) and Flow Matching (FM). DPMs \cite{ldm, ddpm, ddim, iddpm, sde, beats} perform generative modeling by progressively diffusing data into Gaussian noise and learning the reverse denoising process. They have demonstrated exceptional performance in diverse tasks, including unconditional video generation \cite{dit, sora, sd3}, text-to-image synthesis \cite{ldm, dalle2, imagen}, text-to-video generation \cite{sora, animatediff}, and conditional image generation \cite{controlnet, animate_anyone}. FM \cite{cfm, rectflow} has emerged as a powerful alternative, offering faster sampling speeds while achieving sample quality comparable to DPMs \cite{cfm, lcfm, voicebox, boosting_flow_matching, moviegen}. As a flow-based generative model, it estimates a transformation from a prior distribution (e.g., Gaussian) to the target data distribution. Unlike normalizing flows \cite{normalizing_flow, realnvp}, which directly estimate the noise-to-data mapping under specific architectural constraints, FM employs flexible architectures to regress a time-dependent vector field that generates the flow by solving the corresponding ordinary differential equation (ODE) \cite{node}.

\noindent \textbf{Representation Learning as Regularization for Diffusion~}~ In the context of image generation, a recent line of work has explored representation learning as a form of regularization for diffusion models. REPA \cite{repa} explicitly injects external priors by aligning the intermediate representations of the generator with those from a frozen, high-capacity, pre-trained encoder, which can be trained on external data with diverse objectives. Building on this, SARA \cite{SARA} integrates structured and adversarial alignment to enforce multi-level, fine-grained representation consistency. U-REPA \cite{U-REPA} adapts this concept to the diffusion U-Net, mitigating the spatial-semantic mismatch via MLP upsampling and manifold constraints. REPA-E \cite{REPA-E} enables end-to-end joint optimization of a VAE and a Latent Diffusion Model (LDM), introducing a differentiable normalization layer to stabilize statistical propagation between them. For cross-modal text-to-image alignment, SoftREPA \cite{SoftREPA} employs soft text tokens and contrastive learning to enhance semantic consistency and editability. VA-VAE \cite{VA-VAE} reconstructs the latent variable structure from its foundations, significantly accelerating the convergence of subsequent Diffusion Transformers (DiTs). disperse \cite{Disperse}, in turn, enhances the diversity and fidelity of generated outputs by encouraging the dispersion objective of internal representations.

\begin{figure}
    \centering
    \includegraphics[width=\linewidth]{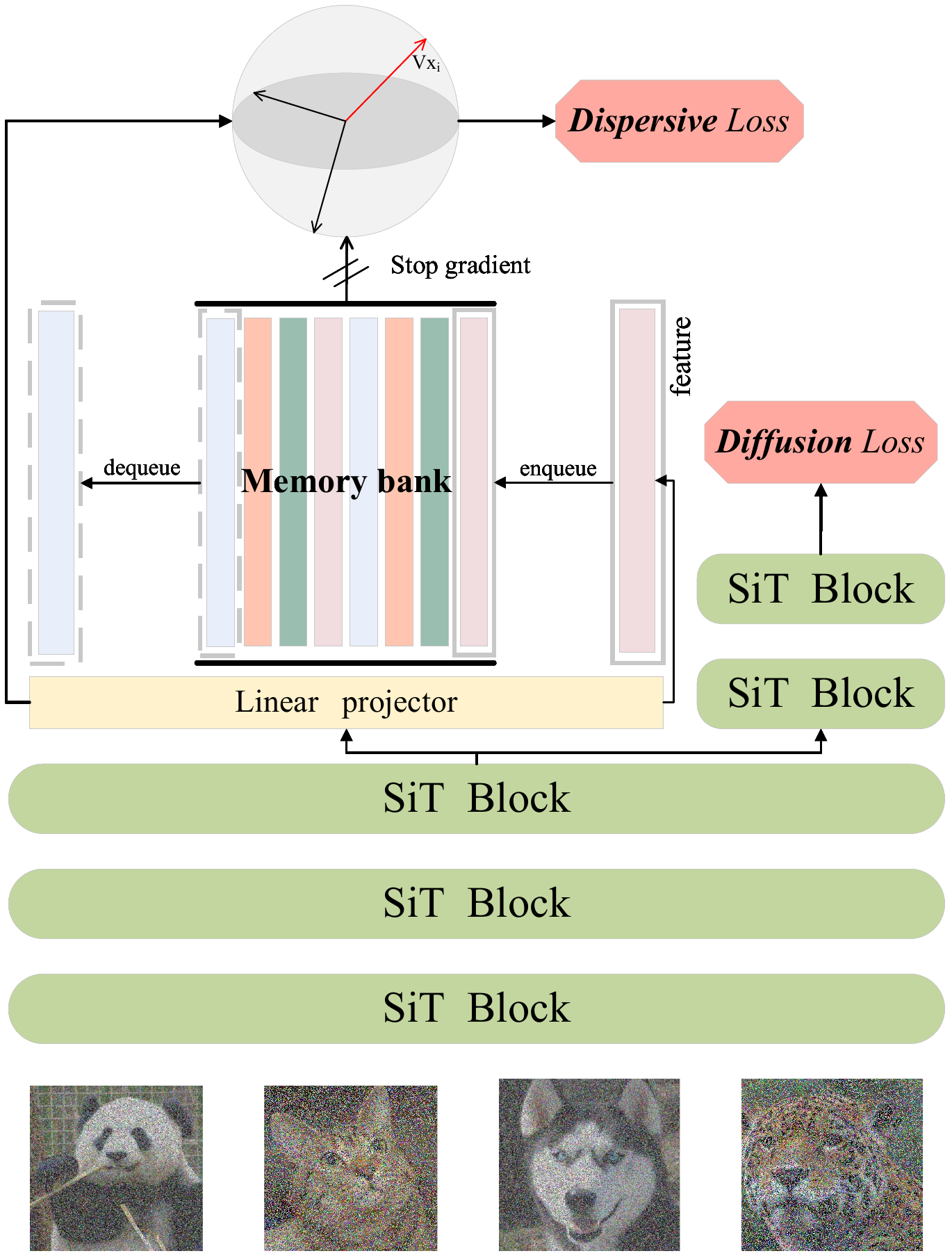}
    \vspace{-10pt}
    \caption{Framework of the proposed {\mname}. We introduce the {\mname} loss into the intermediate layers of denoising-based generative models (e.g., diffusion and flow-matching) to accelerate training convergence. By integrating a memory bank with a low-dimensional projection head, our approach decouples the effective negative sample size from the mini-batch size. This design significantly enriches the diversity of negative samples and strengthens the optimization signal, thereby enhancing feature separability.}
    \label{fig:framework}
\end{figure}

\section{Preliminaries}

\textbf{Contrastive Representation Learning. }Contrastive learning aims to structure an embedding space by attracting semantically related samples, which form positive pairs, and repelling unrelated ones, which constitute negative pairs. Formally, let $\mathbf{z}_i=f_{\theta}(\mathbf{h}_i)$ be the representation of an input $\mathbf{h}_i$, where $f_{\theta}$ denotes the sub-network that computes this representation. Then, the InfoNCE loss can be defined as:

\begin{equation}
\label{eqn:infonce}
    \mathcal{L}_{Cont} = -  \log \frac{e^{\operatorname{sim}(\mathbf{z}_{i},\mathbf{z}_{i}^+)/\tau}}{\sum_{j}e^{\operatorname{sim}(\mathbf{z}_i,\mathbf{v}_j)/ \tau}} 
\end{equation}

where $(\mathbf{z}_{i},\mathbf{z}_{i}^+)$ denotes a positive pair, where $\mathbf{z}_{i}^+$ is an augmented view of $\mathbf{z}_{i}$ obtained through data augmentation. The term $\operatorname{sim}(\mathbf{z}_i, \mathbf{z}_j)=\mathbf{z}_i^\top\cdot \mathbf{z}_j/(\|\mathbf{z}_i\|_2\cdot\|\mathbf{z}_j\|_2)$ represents the cosine similarity between two vectors. $\tau$ is a temperature hyperparameter, which we set to 1 for simplicity.

\textbf{Dispersive Loss~}~The core principle of Dispersive Loss is to enforce mutual repulsion among feature representations within the latent space. This effect is analogous to negative sample repulsion in self-supervised contrastive learning. Diverging from conventional contrastive methods, Dispersive Loss is formulated as a positive-pair-free contrastive objective. This design is predicated on the assumption that the primary generative task, governed by the diffusion objective $\mathcal{L}_{Diff}$, already provides a sufficient alignment signal. Therefore, the Dispersive Loss $\mathcal{L}_{Disp}$ is incorporated as an auxiliary regularization term alongside the standard training objective of the diffusion model. For a given data batch $X=\{x_i\}$, the total loss is defined as:

\begin{equation}
\label{eqn:total_loss}
    \mathcal{L}(X)=\mathbb{E}_{x_{i}\in X}[\mathcal{L}_{Diff}(x_{i})]+\lambda\mathcal{L}_{Disp}(X)
\end{equation}
where the dispersive loss $\mathcal{L}_{Disp}$ is computed over the entire mini-batch, where $\lambda$ is a hyperparameter that controls the trade-off between the regression objective and the regularization strength. Within the InfoNCE framework, the dispersive loss discards the alignment term for positive pairs, exclusively retaining the repulsion term across arbitrary sample pairs, which can be formulated as:

\begin{equation}
\label{eqn:loss_disp}
    \mathcal{L}_{Disp}=log\mathbb{E}_{i,j}[e^{-\operatorname{sim}(z_{i},z_{j})/\tau}]
\end{equation}
where $sim(\cdot,\cdot)$ means the similarity metric function.
% However, under mini-batch training, Dispersive Loss could induce excessive local repulsion, which in turn degrades the diversity of the generated samples.

\begin{figure*}
  \centering
  \includegraphics[width=\textwidth]{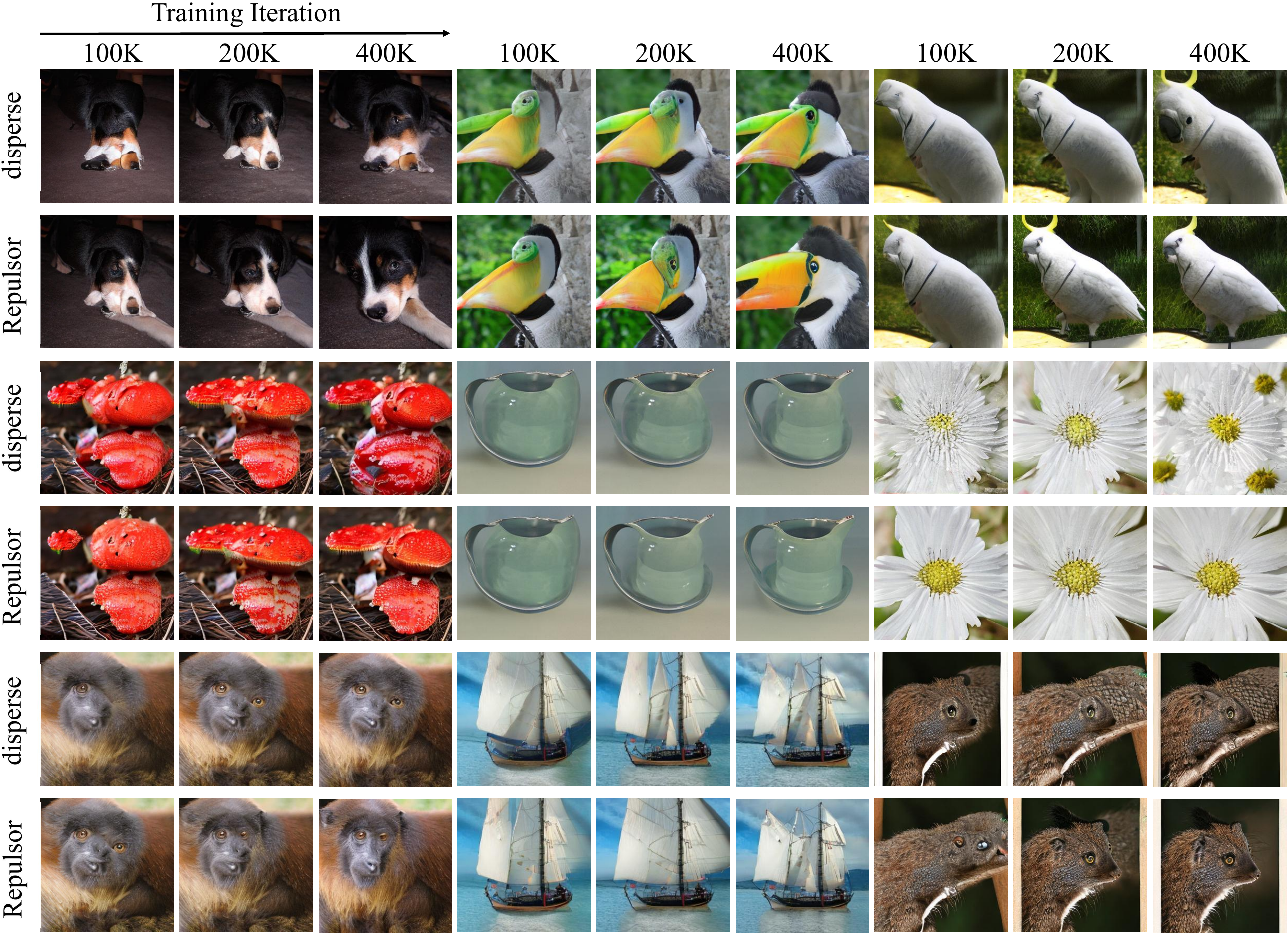}
  \caption{Generated Image Comparison of {\mname} vs disperse~\cite{Disperse}. To evaluate the effect of disperse and {\mname}, we compare images generated by two SiT-XL/2 models during the first 400K training iterations. For a fair comparison, both models are run without classifier-free guidance with $\omega=1.5$ and share the exact same initial noise, sampler, and number of sampling steps.}
  \label{fig:iteration}
\end{figure*}

\section{{\mname}}

Given a mini-batch VAE latent images $\mathbf{X}=\{\mathbf{x}_i\}$, we first sample a diffusion timestep $t \sim \text{Uniform}\{1, \dots, T\}$. Then, the diffusion forward process can be written as: 
% where $T$ is the total number of diffusion timesteps, and subject each sample $\mathbf{x}_0$ to $t$ steps of the forward noising process prior to feature extraction:
\begin{equation}
\label{eqn:diff_xt}
\mathbf{x}_t =\sqrt{\bar{\alpha}_t}\mathbf{x}_0 + \sqrt{1-\bar{\alpha}_t}\boldsymbol{\epsilon},
\quad \boldsymbol{\epsilon}\sim\mathcal{N}(\mathbf{0},\mathbf{I}),\ \bar{\alpha}_t=\prod_{s=1}^{t}(1-\beta_s)
\end{equation}

\begin{algorithm}[t]
\caption{Repulsor training in PyTorch-like style.}
\label{alg:code}
% \algcomment{\fontsize{7.2pt}{0em}\selectfont \texttt{bmm}: batch matrix multiplication; \texttt{mm}: matrix multiplication; \texttt{cat}: concatenation.
% %\vspace{-1.em}
% }
\definecolor{codeblue}{rgb}{0.25,0.5,0.5}
\lstset{
  backgroundcolor=\color{white},
  basicstyle=\fontsize{7.2pt}{7.2pt}\ttfamily\selectfont,
  columns=fullflexible,
  breaklines=true,
  captionpos=b,
  commentstyle=\fontsize{7.2pt}{7.2pt}\color{codeblue},
  keywordstyle=\fontsize{7.2pt}{7.2pt},
%  frame=tb,
}
\begin{lstlisting}[language=python]
# f: encoder networks
# queue: dictionary as a queue of K keys (CxK)
# tgt: target layer for disperse loss
# m: momentum
# t: temperature
# gamma: loss weight

opt = AdamW(f.parameters()) # # Setup optimizer

# load a minibatch x and label y with N samples
for x, y in loader:
    x_n = VAE(x) # to latent space + normalize latents
    # iterate over f layers
    for i in range(len(self.blocks)): 
        # choose one layer
        q = f.blocks[i](x_n)  
        # target layer tgt for computing l_disp
        if i == tgt:
            # linear projection of q
            k = linear_proj(q.view(q.size(0), -1))
            
            # enqueue the current minibatch
            enqueue(queue, k) 
            # dequeue the earliest minibatch
            dequeue(queue)  
    
            # pairwise squared Euclidean distances
            dist = cdist(k, queue.T.detach(), p=2) ** 2
            # disperse loss, Eqn.(7)
            l_disp = log(mean(exp(-dist / t)))

    # compute the conditioned model prediction
    y_pred = f.forward(x_n, y)
    # diffusion loss
    l_diff = mean((y_pred - y) ** 2)
    # total loss calculation
    loss = l_diff + gamma * l_disp

    # SGD update
    opt.zero_grad()
    loss.backward()
    opt.step()
\end{lstlisting}
\end{algorithm}
%##################################################################################################

where $\mathbf{x}_{0}$ is the clean latent and $\mathbf{x}_t$ is the noised latent. $\alpha$ and $\beta$ are pre-defined hyper-parameters. Then the corresponding timestep $t$ and noisy latent $\mathbf{x}_{t}$ will be fed into SiT models to yield intermediate features $\mathbf{H}_t=\{\mathbf{h}_i\}_{i=1}^B$:
\begin{equation}
\label{eqn:h_noise}
\mathbf{h}_b = f_{SiT[1:F_l]}(\mathbf{x}_t,e_t)
\end{equation}
where $f_{SiT[1:F_l]}$ means the first $F_l$ blocks of SiT model. Subsequently, a single-layer linear projection head, $g_{\theta}$, is employed to map the high-dimensional features $\mathbf{h}_b$ into a $D$-dimensional latent space, producing the final latent representations $\mathbf{Z}_t=\{\mathbf{z}_i\}_{i=1}^B$, which can be written as:
\begin{equation}
\label{eqn:linear}
\mathbf{z}_i=\operatorname{norm}(g_{\theta}(\mathbf{h}_i))\in\mathbb{R}^{D}
\end{equation}
where \(\operatorname{norm}(\cdot)\) denotes the \(L_2\) normalization. To increase the number of negatie pairs, we construct a memory bank $\mathcal{M}=\{\mathbf{m}_i\}_{i=1}^{K}$ as a queue of size $K$, where $\mathbf{m}_i \in \mathbb{R}^{1\times D}$ is the $i$-th negative sample. Similar to Moco~\cite{he2020momentum}, the memory bank is updated by the First-In, First-Out policy. Then, the dispersive objective of current batch can be written as:

% A stop-gradient operation is applied to the features within the memory bank to halt gradient backpropagation, thereby minimizing computational and storage costs. Specifically, for the current batch of contrastive features $\mathbf{Z}_t$ and a memory bank $\mathcal{M}_{t-1}$ of capacity $K$, the update proceeds by enqueuing $\mathbf{Z}_t$ and simultaneously dequeuing the oldest batch of features. This procedure maintains a constant bank size, $|\mathcal{M}_t|=K$, and produces the updated memory bank $\mathcal{M}_{t}$.

% During training, we enforce a dispersion constraint on the feature representations. For each anchor $\mathbf{z}_i \in \mathbf{Z}_t$, its negative set is all entries in the current memory bank $\mathcal{M}_{t}$. The objective is to maximize the squared Euclidean distance $D_{ij}$ to these negative samples, thereby promoting feature diversity across the embedding space. This is achieved by minimizing the following disperse loss $\mathcal{L}_{\mathrm{Disp}}$:

\begin{equation}
\label{eqn:new_disp}
\mathcal{L}_{\mathrm{Disp}} =\log \frac{1}{BK}\sum_{i=1}^{B}\sum_{k=1}^{K}
\exp\left(-\frac{D_{ik}}{\tau}\right) 
\end{equation}
where $D_{ik}= \|\mathbf{z}_i - \operatorname{sg}(\mathbf{m}_k)\|_2^2$ is defined as the squared Euclidean distance and $\operatorname{sg}(\cdot)$ denotes the stop-gradient operation. After calculating the dispersive loss, we fed the latent representation $\mathbf{Z}_{t}$ into the remaining part of SiT model $f_{SiT[F_l:F_f]}$, where $F_f$ is total the number of blocks of SiT model to yeild noise prediction $\hat{\mathbf{\epsilon}}$ to calculate diffusion loss.

Finally, we formulate the overall training objective as a weighted sum of the diffusion loss and the disperse loss:
\begin{equation}
\label{eqn:total_loss_ours}
\mathcal{L} = \mathcal{L}_{Diff} + \gamma \cdot \mathcal{L}_{Disp}
\end{equation}
where $\gamma$ is a hyperparameter that balances the disperse and diffusion loss. The memory bank and the projection head are introduced as auxiliary modules exclusively for the optimization process during training. At inference time, these modules are detached, ensuring the efficiency of {\mname}. Algorithm \ref{alg:code} presents the pre-training pseudocode.

\section{Experimental}
% \begin{table}[tb!]
% 	\centering
% 	% \footnotesize
%     % \small
%     \caption{Configurations of SiT architecture at different model sizes. }
% 	\setlength{\belowcaptionskip}{0cm}
% 	\begin{tabular}{ccccc}
% 		\toprule
% 		Model & Params (M) & Depths  & Channel & \# Heads  \\
% 		\midrule
% 		\textbf{SiT-B/2} & 130 & 12  & 768 & 12  \\
% 		\textbf{SiT-XL/2} & 675 & 28  & 1152 & 16  \\
%         \midrule
%         Model & Epochs & Batch  & {$\gamma$} & {$\tau$}  \\
%         \textbf{SiT-B/2} & 80 & 256  & 0.25 & 0.5  \\
% 		\textbf{SiT-XL/2} & 80-200 & 256  & 0.25 & 0.5  \\
% 		\bottomrule
% 	\end{tabular}
% 	% \vspace{-5pt}
% 	\label{tab:configurations}
%   % \vspace{-10pt}
% \end{table}
\textbf{Implementation details.~}~Our experiments adopt the SiT \cite{sit} architecture, and the original model serves as our primary baseline. To ensure a fair comparison, we strictly adhere to the implementation settings from \cite{sit,Disperse} and conduct our evaluations on ImageNet \cite{imagenet} at a fixed $256\times256$ resolution. The training is performed in a $32\times32\times4$ latent space encoded by a VAE tokenizer \cite{ldm}. For optimization, we employ the AdamW optimizer \cite{adamw} with a learning rate of $1\times10^{-4}$, $(\beta_1,\beta_2)=(0.9,0.95)$, and with no weight decay. During the sampling phase, we use an ODE-based Heun sampler with 250 fixed steps, consistent with \cite{Disperse,sit}. All experiments were conducted on 8 A100 (80GB) GPUs. The configuration details can be found in Appendix \ref{App:Additional Experimental Details}.

\textbf{Comparison methods.~}~In experiments evaluating {\mname} under CFG, we benchmark against a diverse suite of recent diffusion-based generative models that vary in input parameterizations and architectural design. Specifically, we consider four families: (a) pixel-space diffusion, including ADM-U \cite{dhariwal}, VDM$++$ \cite{vdmpp}, and Simple Diffusion \cite{simplediffusion}; (b) Transformer-based latent diffusion, covering U‑ViT \cite{uvit}, DiffiT \cite{diffit}, DiT \cite{dit}, and SiT \cite{sit}; (c) masked diffusion transformers, represented by MaskDiT \cite{maskdit} and MDTv2 \cite{mdt}; and (d) approaches that leverage external pretrained encoders for representation alignment, including REPA \cite{repa}, U‑REPA \cite{U-REPA}, and SARA \cite{SARA}.

\vspace{-1pt}
\subsection{Main Results}
% \vspace{-1pt}

We present a comprehensive evaluation of various SiT models trained with {\mname}. All models, except SiT-XL/2, were trained for 400k iterations. Across all experiments, we uniformly set the hyperparameters to $\gamma=0.25$ and $\tau=0.5$. For SiT-XL/2, we comparatively analyze its performance with and without Classifier-Free Guidance (CFG), providing both quantitative evaluations and a qualitative assessment of the generated samples. In the following experiments, our results are compared against the SiT performance reported in disperse \cite{Disperse}.

\begin{table}[tb!]
  \centering
  \setlength{\belowcaptionskip}{0cm} 
    \caption{\textbf{Performance comparison of {\mname}, disperse, and SiT under the classifier-free guidance-free setting.} The results show that SiT integrated with {\mname} achieves the best performance. $\dagger$: applied to the single best block. $\ddagger$: applied to all blocks.}
    \vspace{-8pt}
  \resizebox{\linewidth}{!}{%
\begin{tabular}{lccc}
  \toprule
  \multicolumn{4}{l}{\bf{ImageNet} 256$\times$256, w/o cfg} \\
  \toprule
  Model  & Iter. & FID$\downarrow$ & \textbf{$\Delta$} \\
  \midrule
  SiT-S/2  & 400K & 60.63 & -\\
  SiT-S/2+dispersive  & 400K & 58.45 & \textcolor{Green}{--2.18\,(--3.60\%)}\\
  \rowcolor{gray} SiT-S/2+{\mname}  & 400K & \textbf{54.36} & \textcolor{Green}{--6.27\,(--10.34\%)}\\
  \midrule
  SiT-B/2  & 400K & 36.49 & -\\
  SiT-B/2+dispersive$^\dagger$  & 400K & 32.35 & \textcolor{Green}{--4.14\,(--11.35\%)}\\
  SiT-B/2+dispersive$\ddagger$  & 400K & 32.05 & \textcolor{Green}{--4.44\,(--12.17\%)}\\
  \rowcolor{gray} SiT-B/2+{\mname}  & 400K & \textbf{27.46} & \textcolor{Green}{--9.03\,(--24.75\%)}\\
  \midrule
  SiT-L/2  & 400K & 20.41 & -\\
  SiT-L/2+dispersive  & 400K & 16.68 & \textcolor{Green}{--3.73\,(--18.27\%)}\\
  \rowcolor{gray} SiT-L/2+{\mname}  & 400K & \textbf{16.43} & \textcolor{Green}{--3.98\,(--19.50\%)}\\
  \midrule
  SiT-XL/2 & 400K & 18.46 & - \\
  SiT-XL/2+dispersive & 400K & 15.95 & \textcolor{Green}{--2.51\,(--13.6\%)}\\
  \rowcolor{gray} SiT-XL/2+{\mname}  & 400K & \textbf{10.91} & \textcolor{Green}{--7.55\,(--40.9\%)} \\
  \midrule
  SiT-XL/2 & 700K & 14.06 & - \\
  SiT-XL/2+dispersive & 700K & 12.08 & \textcolor{Green}{--1.98\,(--14.1\%)} \\
  \rowcolor{gray} SiT-XL/2+{\mname}  & 700K & \textbf{9.93} & \textcolor{Green}{--4.13\,(--29.4\%)} \\
  \midrule
  SiT-XL/2 & 1000K & 12.18 & - \\
  SiT-XL/2 & 2000K & 10.11 & - \\
  SiT-XL/2+dispersive & 1000K & 10.64 & \textcolor{Green}{--1.54\,(--12.6\%)} \\
  \rowcolor{gray} SiT-XL/2+{\mname}  & 1000K & \textbf{9.07} & \textcolor{Green}{--3.11\,(--25.5\%)} \\
  \bottomrule
  \end{tabular}
  }
  \vspace{5pt}
  \label{tab:nocfg}
  % \vspace{-10pt}
\end{table}

\begin{table}[tb!]
  \centering
  \setlength{\belowcaptionskip}{0cm}  
  \caption{\textbf{FID comparisons based on classifier-free guidance using SiT-XL/2 backbone.} {\mname} achieves comparable performance in only 80 epochs when compared to the 400 epochs for mini-batch disperse~\cite{Disperse}, yielding an approximately \textbf{5$\times$} speed-up in convergence while attaining a slightly lower FID.}
  \vspace{-8pt}
  % \resizebox{\textwidth}{!}{%
\begin{tabular}{lcc}
  \toprule
  \multicolumn{3}{l}{\bf{ImageNet} 256$\times$256, w/ cfg} \\
  \toprule
  Model  & Epochs & FID$\downarrow$ \\
  \midrule
  \emph{Pixel diffusion} & & \\
  ADM-U~\cite{dhariwal}      & 400  & 3.94 \\
  VDM$++$~\cite{vdmpp}   & 560  & 2.40 \\
  Simple diffusion~\cite{simplediffusion} & 800  & 2.77 \\

  \midrule
  % \multicolumn{3}{l}{\emph{Latent Diffusion Transformer}\vspace{0.02in}} \\
  \emph{Latent Diffusion Transformer} & & \\
     U-ViT-H/2~\cite{uvit}  & 240  & 2.29 \\ 
     DiffiT~\cite{diffit}  & -    & 1.73 \\
     DiT-XL/2~\cite{dit}   & 1400 & 2.27 \\
     % SiT-XL/2~\cite{sit}   & 80 & 6.02 \\
     % SiT-XL/2~\cite{sit}   & 140 & 3.95 \\
     SiT-XL/2~\cite{sit}   & 200 & 3.30 \\
     SiT-XL/2~\cite{sit}   & 800 & 2.46 \\
     SiT-XL/2~\cite{sit}   & 1400 & 2.06 \\

  \midrule
  % \multicolumn{3}{l}{\emph{Masked Diffusion Transformer}} \\
  \emph{Masked Diffusion Transformer} & & \\
     MaskDiT~\cite{maskdit} & 1600 & 2.28 \\ 
     MDTv2-XL/2~\cite{mdt} & 1080 & 1.58 \\

  \midrule
  % \multicolumn{3}{l}{\emph{Representation Alignment}} \\
  \emph{Representation Alignment } & & \\
     SiT-XL/2 + REPA~\cite{repa}       & 800  & 1.42 \\
     SiT-XL/2 + U-REPA~\cite{U-REPA} & 400  & 1.41 \\
     SiT-XL/2 + SARA~\cite{SARA} & 400  & 1.36 \\
  \midrule
  % \multicolumn{3}{l}{\emph{Dispersion Regularization}} \\
  \emph{Dispersion Regularization } & & \\
    SiT-XL/2 +disperse~\cite{Disperse} & 40 & 7.32 \\
    SiT-XL/2 +disperse~\cite{Disperse} & 80 & 5.09 \\
    SiT-XL/2 +disperse~\cite{Disperse} & 140 & 3.42 \\
    \rowcolor{gray} SiT-XL/2 +{\mname} (Ours) & 40 & \textbf{3.25} \\
    % \rowcolor{gray} SiT-XL/2 +{\mname} (Ours) & 140 & \textbf{2.47} \\
    SiT-XL/2 +disperse~\cite{Disperse} & 200 & 2.90 \\
    \rowcolor{gray} SiT-XL/2 +{\mname} (Ours) & 80 & \textbf{2.40} \\
    % SiT-XL/2 +disperse~\cite{Disperse} & 400 & 2.39 \\
    % \rowcolor{gray} SiT-XL/2 +{\mname} (Ours) & 200 & \textbf{2.60} \\
  \bottomrule
  \end{tabular}
  % }
  \vspace{5pt}
  \label{tab:cfg}
  % \vspace{-10pt}
\end{table}

\textbf{w/o CFG.~}~As detailed in Table \ref{tab:nocfg}, in the absence of CFG , our proposed method, {\mname}, consistently outperforms disperse by achieving a lower Fréchet Inception Distance (FID; \cite{heusel2017gans}) score throughout the entire training process on the SiT-XL/2 configuration. Notably, at just 400k iterations, {\mname} achieves an FID of 10.91, a score that is not only superior to disperse at 700k iterations (FID=12.08) but also surpasses the SiT baseline after 1M iterations. This represents a substantial improvement in training efficiency, reducing computational time by over 60\%. Furthermore, Figure \ref{fig:iteration} presents a qualitative comparison at different training stages, initialized with the same noise vector. The results visually demonstrate that the model trained with {\mname} exhibits a more favorable evolutionary trajectory, consistently generating higher-quality images at all evaluated stages.

\textbf{w/ CFG.~}~We further conduct a quantitative evaluation of the SiT-XL/2 architecture under CFG. Our approach is benchmarked against several recent representative diffusion models across multiple metrics. To ensure a fair comparison, we adhere to the experimental setup of disperse and SiT, employing a fixed guidance scale of $w=1.5$ without any hyperparameter fine-tuning. As presented in Table \ref{tab:cfg}, {\mname} demonstrates remarkable training efficiency. It achieves performance comparable to disperse trained for 400 epochs with only 80 training epochs, a five-fold reduction in training cost. Moreover, {\mname} surpasses the original SiT-XL/2 with only one-tenth the training epochs.

\vspace{-1pt}
\subsection{Ablation Study}
% \vspace{-1pt}

\begin{figure*}
  \centering
  \includegraphics[width=\textwidth]{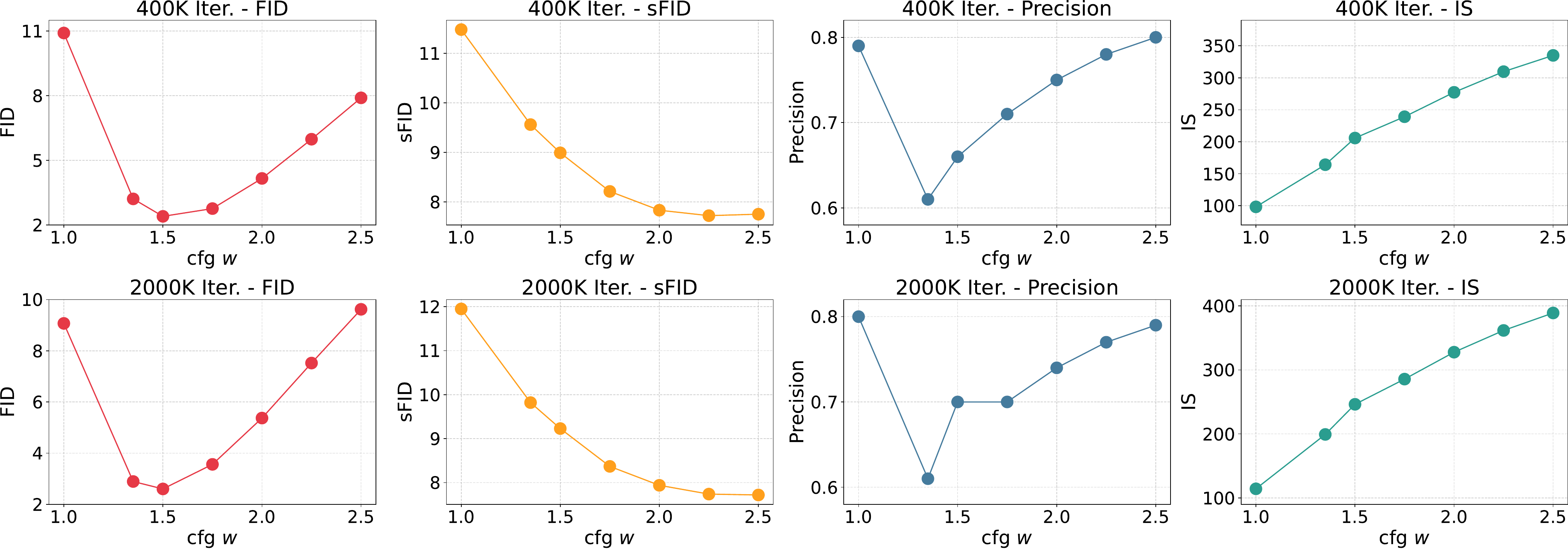}
  \vspace{-10pt}
  \caption{Impact of the CFG ratio $w$ on generation quality. The eight subplots illustrate four metrics (FID$\downarrow$, sFID$\downarrow$, Precision$\uparrow$, and IS$\uparrow$) as a function of $w$ at two different training checkpoints, namely 400K and 2000K iterations.}
  \label{fig:cfgratio}
\end{figure*}

\begin{table}[tb!]
    \centering
      \centering
      % \small
      \caption{\textbf{Ablation Study on Regularization Block Placement on SiT-B/2.}
We apply {\mname} to different blocks and measure the FID. The results clearly demonstrate that the proposed loss significantly improves generation quality in all tested settings. $\dagger$: applied to the single best block. $\ddagger$: applied to all blocks.
}\label{depth}
\vspace{-8pt}
      \begin{tabular}{lcc}
        \toprule
        {Model} & {FID$\downarrow$}  & $\Delta$\\
        \midrule
        SiT-B/2 & 36.49 & -\\
        SiT-B/2+dispersive$\dagger$  & 32.35 & \textcolor{Green}{--4.14\,(--11.35\%)}\\
        SiT-B/2+dispersive$\ddagger$ & 32.05 & \textcolor{Green}{--4.44\,(--12.17\%)}\\
        % \rowcolor{gray} SiT-B/2+{\mname} (Ours) & \textbf{27.46} & \textcolor{Green}{--9.03\,(--24.75\%)}\\
        % \midrule
        % Regularization Block Placement
        % \multicolumn{3}{c}{{\small $Regularization \ Block \ Placement$}} \\
        \midrule
        block 6 & 33.37 & \textcolor{Green}{--3.12\,(--8.55\%)} \\
        block 7 & 32.20 & \textcolor{Green}{--4.29\,(--11.76\%)} \\
        block 8 & \textbf{27.46} & \textcolor{Green}{--9.03\,(--24.75\%)} \\
        block 9 & 30.85 & \textcolor{Green}{--5.64\,(--15.46\%)} \\
        block 10 & 29.55 & \textcolor{Green}{--6.94\,(--19.02\%)} \\
        block 11 & 29.79 & \textcolor{Green}{--6.70\,(--18.36\%)} \\
        block 12 & 34.21 & \textcolor{Green}{--2.28\,(--6.25\%)} \\
        \bottomrule
      \end{tabular}
      % \vspace{.75em}
\end{table}

\textbf{Effect of Regularization Block. }
\textbf{Setups.} For the SiT-B/2 architecture, we apply the {\mname} regularization to only a single Transformer block, while all other training and evaluation configurations are kept identical to the baseline.
\textbf{Results.} As demonstrated in Table \ref{depth}, {\mname} yields significant performance improvements across all placement positions. Remarkably, applying {\mname} to just a single block alone surpasses the performance of applying the Dispersive regularizer to all blocks. This finding further corroborates the superior regularization capability of {\mname} and its ability to achieve higher generation quality.

\begin{table}[t]
    \centering
      \centering
      % \small
      \caption{FID scores of SiT-B/2 on ImageNet with varying downsampling types. ``Use projection'' means whether to use projector in the latent space. ``\#Arch'' means the architecture of projection. ``Avg Pool to $D$'' means performing spatial pooling to $D$ dimension. ``Lin. to $D$'' means concatenating all patches to obtain $1\times PD$ tensor, followed by a linear layer to obtain $1\times D$ token.}\label{linear}
      \vspace{-8pt}
      \resizebox{\linewidth}{!}{\begin{tabular}{lccc}
        \toprule
        Method & Use projection? & \# Arch. & FID$\downarrow$ \\
        \midrule
        SiT~\cite{sit} & \XSolidBrush & - & 36.49\\
        {\mname} & \Checkmark & Avg Pool to $D$ & 36.70\\
        \rowcolor{gray} {\mname} (Ours) & \Checkmark & Lin. to $D$ & \textbf{32.88}\\
        \bottomrule
      \end{tabular}}
      \vspace{.75em}
\end{table}

% \begin{figure*}
%   \centering
%   \includegraphics[width=\textwidth]{figures/mic.pdf}
%   \vspace{-10pt}
%   \caption{Impact of the CFG ratio $w$ on generation quality. The eight subplots illustrate four metrics (FID$\downarrow$, sFID$\downarrow$, Precision$\uparrow$, and IS$\uparrow$) as a function of $w$ at two different training checkpoints, namely 400K and 2000K iterations.}
%   \label{fig:cfgratio}
% \end{figure*}

\textbf{Effect of Linear Projection.} 
\textbf{Setups.} \textbf{Setups.} Using the SiT-B/2 backbone, we compare the SiT~\cite{sit} (no projector) against two projector architectures: ``Avg Pool to $D$'' and our proposed ``Lin. to $D$'', as detailed in Table \ref{linear}. All other configurations remain consistent.
\textbf{Results.} As shown in Table \ref{linear}, the SiT baseline (FID 36.49) suffers from overfitting when using high-dimensional representations, and the ``Avg Pool to $D$'' (FID 36.70) offers no improvement. Our proposed ``Lin. to $D$'' projector, however, creates an effective information bottleneck that suppresses noise and reduces overhead. This method achieves the best FID of 32.88, significantly outperforming both alternatives.

\textbf{Effect of CFG ratio.}
\textbf{Setups}. We analyze the effect of the CFG ratio $w$ on generation quality on SiT-XL/2 under two distinct training budgets, 400k and 2000k steps. The evaluation is conducted using four standard metrics: FID, spatial FID (sFID; \cite{nash2021generating}), Precision \cite{kynkaanniemi2019improved}, and Inception Score (IS; \cite{salimans2016improved}). We provide more details of each metric in Appendix \ref{appen:eval}. \textbf{Results}. We observe that the FID score exhibits a characteristic U-shaped curve as a function of $w$, consistently reaching its optimum at approximately $w=1.5$. In contrast, sFID shows a monotonic decrease with increasing $w$, while the Inception Score monotonically increases. Precision also demonstrates an overall improvement trend, indicating that stronger guidance yields sharper samples with higher class consistency. Mechanistically, a larger $w$ amplifies the differential between the conditional and unconditional paths, steering the generation more closely toward the condition, yet simultaneously elevating the risk of mode collapse and distributional shift.

\begin{table}[t]
    \centering
      \centering
      % \small
      \caption{FID scores of SiT-B/2 on ImageNet with varying memory bank sizes. ``no memory bank'' means treating samples in the same mini-batch as negatives. ``Num. Negatives'' means the number of negative samples, i.e., memory bank size $K$.
}\label{memory}
\vspace{-8pt}
\resizebox{\linewidth}{!}{%
      \begin{tabular}{lcccc}
        \toprule
        Method & Batch Size & Memory Bank & Num. Negatives & FID$\downarrow$ \\
        \midrule
        disperse~\cite{Disperse} & 256 & \XSolidBrush & - & 32.88\\
        {\mname} (Ours) & 256 & \Checkmark & 65536 & 32.64\\
        {\mname} (Ours) & 256 & \Checkmark & 131072 & \textbf{27.46}\\
        {\mname} (Ours) & 256 & \Checkmark & 262144 & 34.79\\
        \bottomrule
      \end{tabular}
      }
      % \vspace{.75em}
\end{table}

\textbf{Effect of Memory Bank Size.} 
\textbf{Setups.} We investigate the impact of varying memory bank sizes, denoted by $K$, on the SiT-B/2 architecture. We also include \textbf{disperse}, which serves as a "no memory bank" ablation (i.e., $K=0$) where negative samples are drawn exclusively from the current mini-batch. All other training configurations remain consistent with our main experiments.
\textbf{Results.} As presented in Table \ref{memory}, the memory bank size significantly influences performance. The optimal FID score of 27.46 is achieved with $K=131072$, outperforming both disperse and other memory configurations. An excessively large memory bank can dilute the contribution of in-batch hard negatives, reducing the gradient's sensitivity to informative samples, which in turn weakens regularization and leads to performance degradation. Conversely, a memory bank that is too small or absent provides an insufficient approximation of the true data distribution, limiting generalization and failing to maintain a stable dispersion intensity.

% The size of the memory bank is a critical hyperparameter. As detailed in Table \ref{memory}, we determined an optimal size of 131072 for our experiments with SiT-B/2. An excessively large memory bank can be counterproductive, as it dilutes the contribution of hard negatives within the batch mean. This diminishes the gradient's sensitivity to these informative samples, thereby attenuating the regularization impetus. Conversely, a memory bank that is too small provides a poor approximation of the true data distribution, which hinders generalization. It also struggles to maintain a stable dispersion intensity, limiting the overall effectiveness of the regularization.

\begin{figure}[t]
    \centering
    % \label{fig:sit}
    \begin{subfigure}[b]{0.49\linewidth}
        \centering
        \includegraphics[width=\linewidth]{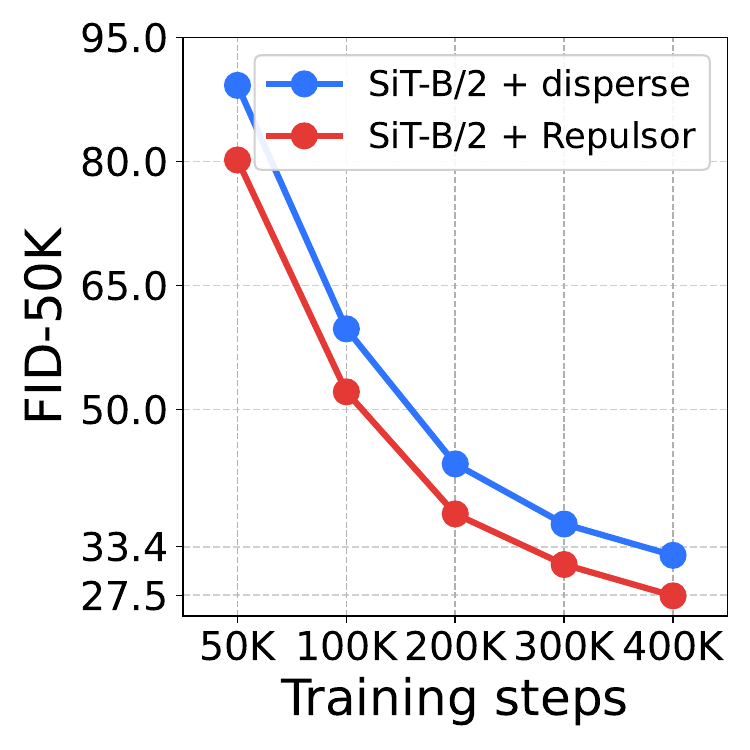}
        \label{fig:sitb}
    \end{subfigure}
    \hfill
    \begin{subfigure}[b]{0.49\linewidth}
        \centering
        \includegraphics[width=\linewidth]{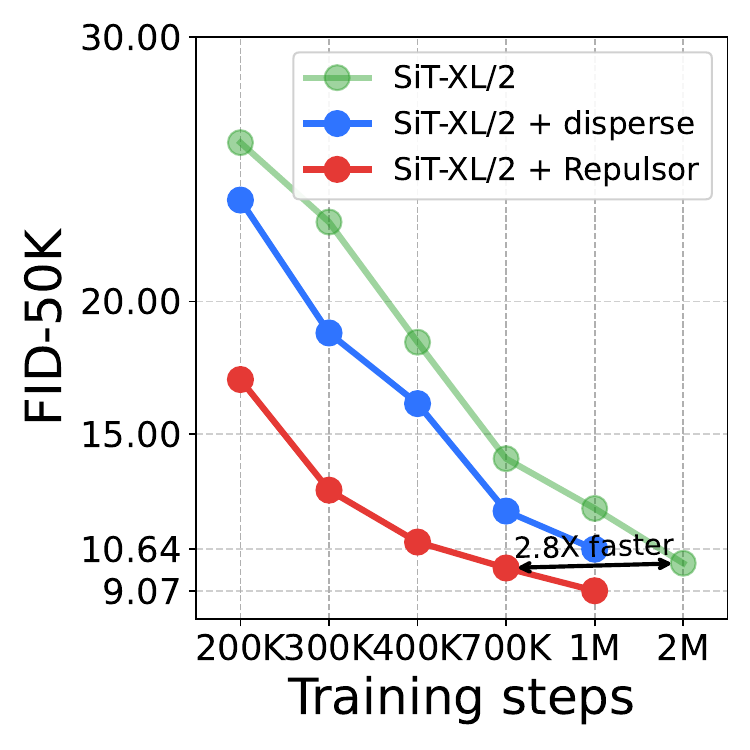}
        \label{fig:sitxl}
    \end{subfigure}
    \vspace{-8pt}
    \vspace{-10pt}
    \caption{\textbf{Effectiveness of {\mname} on SiT with different architectures (SiT-B/2 on the left and SiT-XL/2 on the right).} We plot the FID-50k score on the ImageNet 256$\times$256 dataset as a function of the training iterations. Compared to disperse, {\mname} consistently attains a lower FID score with considerably fewer iterations, demonstrating a significant improvement in both the training efficiency and generation quality of diffusion models.}
    \label{fig:sit_comparison}
\end{figure}

% \begin{figure}
%     \centering
%     \includegraphics[width=.8\linewidth]{figures/sitb.pdf}
%     \caption{\textbf{Effectiveness of {\mname} on SiT-B/2.} We plot the FID-50k score on ImageNet 256×256 as a function of training iterations. Compared to disperse, {\mname} attains a lower FID with considerably fewer iterations, demonstrating a significant improvement in both the training efficiency and generation quality of diffusion models.}
%     \label{fig:}
% \end{figure}

% \begin{figure}
%     \centering
%     \includegraphics[width=.8\linewidth]{figures/sitxl.pdf}
%     \caption{\textbf{Effectiveness of {\mname} on SiT-XL/2.} {\mname} yields a 2.8x convergence speedup over the baseline, markedly accelerating the training process.}
%     \label{fig:}
% \end{figure}

\textbf{Effect of Training Steps.} \textbf{Setups.} We compare {\mname} against the baseline on two SiT architectures (SiT-B/2 and SiT-XL/2).
\textbf{Results.} As shown in Figure \ref{fig:sit_comparison}, on the SiT-B/2 architecture, {\mname} consistently achieves a lower FID throughout the training process, yielding an average relative improvement of approximately 13.1\%. On the larger SiT-XL/2 architecture, the advantages of {\mname} become more pronounced. To achieve an FID score below 11, {\mname} requires only 400K training steps (FID=10.91), whereas disperse necessitates 1M steps (FID=10.64). Furthermore, {\mname} demonstrates significantly enhanced sample efficiency. To achieve a comparable FID score, {\mname} requires only 700K steps (FID=9.93), whereas the original SiT model needs approximately 2M steps (FID=10.11). This demonstrates that {\mname} enhances sample efficiency by nearly 2.8$\times$. When pushing performance further, {\mname} attains an even stronger score of 9.07 within just 1M steps. These results indicate that the performance gains from {\mname} are particularly significant during the early-to-mid stages of training and are further amplified on larger-scale models.

\begin{table}[t]
    \centering
      \centering
      % \small
      \caption{Sensitivity analysis on $\tau$ and $\gamma$. We evaluate the performance of SiT-XL/2 on {\mname} with varying temperature $\tau$ and loss weight $\gamma$. All configurations outperform disperse.
}\label{tab:coeff}
\vspace{-8pt}
\resizebox{\linewidth}{!}{%
      \begin{tabular}{lcccc}
    \toprule
    Method & Temperature  $\tau$ & Loss weight $\gamma$ & FID$\downarrow$ & Inception score$\uparrow$\\
    \midrule
        SiT-XL/2~\cite{sit} & - & - & 18.46 & 73.9 \\
        % {\mname} & - & - & 10.91 & 98.3 \\
    \midrule
    
        \multirow{3}{*}{{\mname}} & 0.25 & 0.25 &  12.49 & 88.1 \\
        ~ & 0.25 & 0.5  & 11.40 & 88.9 \\
        ~ & 0.5 & 0.25  & \textbf{10.91} & \textbf{98.3} \\
       %0.5 & 0.5   & 23.15 & 58.8 \\
    \bottomrule
  \end{tabular}
      }
      % \vspace{.75em}
\end{table}

% \begin{table}
%   \centering
%   \caption{Sensitivity analysis on $\tau$ and $\gamma$. We evaluate the performance of SiT-XL/2 on {\mname} with varying temperature $\tau$ and loss weight $\gamma$. All configurations outperform disperse.}
%   \vspace{-8pt}
%   \resizebox{\linewidth}{!}{
%   \begin{tabular}{cccc}
%     \toprule
%     $\tau$ & $\gamma$ & FID$\downarrow$ & IS$\uparrow$\\ \\
%     \midrule
%     0.25 & 0.25 &  12.49 & 88.1 \\
%         0.25 & 0.5  & 11.40 & 88.9 \\
%         0.5 & 0.25  & 10.91 & 98.3 \\
%        %0.5 & 0.5   & 23.15 & 58.8 \\
%     \bottomrule
%   \end{tabular}}
%   \label{tab:coeff}
% \end{table}

\textbf{Effect of $\tau$, $\gamma$.} \textbf{Setup. }We further investigate the two key hyperparameters in our loss function: the temperature $\tau$ and the regularization strength $\gamma$. We employ SiT-XL/2 and examine the impact of varying $\tau$ and $\gamma$ on performance under the standard 400k-iteration training setting.
\textbf{Results.} The results are presented in Table \ref{tab:coeff}. We observe that the model achieves optimal performance with $\tau=0.5$ and $\gamma=0.25$. While an exhaustive grid search was computationally prohibitive, this parameter combination demonstrated robust and optimal performance within our explored range.

% \begin{table}[h]
%   \centering
%   % \small
%   \caption{\textbf{Inception Scores on ImageNet}.}
%   \begin{tabular}{@{}c|cccc|cccc}
%   \toprule
%   &\multicolumn{3}{c}{400K Iter.} & \multicolumn{3}{c}{2000K Iter.} \\
%   \cmidrule(lr){2-4} \cmidrule(lr){5-7}
%   {$w$} & FID$\downarrow$ & sFID$\downarrow$ & Pre.$\downarrow$ & IS$\uparrow$ & FID$\downarrow$ & sFID$\downarrow$  & Pre.$\downarrow$ & IS$\uparrow$\\
%   \midrule
%     1.35 & 3.21 & 9.56 & 0.61 & 164.2 & 2.89 & 9.82 & 0.61 & 199.2\\
%     1.5 & 2.40 & 8.99 & 0.66 & 205.9 & 2.60 & 9.23 & 0.70 & 246.3\\
%     1.75 & 2.76 & 8.21 & 0.71 & 239.2 & 3.56 & 8.37 & 0.70 & 285.7 \\
%     2.0 & 4.16 & 7.83 & 0.75 & 277.4 & 5.37 & 7.94 & 0.74 & 327.8 \\
%     2.25 & 5.98 & 7.72 & 0.78 & 309.7 & 7.52 & 7.74&  0.77 &361.8 \\
%     2.5 & 7.90 & 7.75 & 0.80 & 335.1 & 9.62 & 7.72 & 0.79 & 389.1\\
%   \bottomrule
% \end{tabular}
% \vspace{.75em}
%   \label{inception_score}
% \end{table}

\section{Conclusion}

We present {\mname}, a simple yet effective regularization method designed to disperse the internal representations of a model. We investigate whether the size of the negative set in generative models follows a scaling law and evaluate its impact on representation quality. Experiments demonstrate that {\mname} significantly improves generative performance and accelerates convergence by dynamically maintaining a memory-efficient queue of negative samples, without requiring any additional data or pre-training. Our work provides new insights and strong empirical evidence for dispersion regularization in generative models.

\textbf{Limitation. }While {\mname} significantly enhances the representation quality of generative models, its application to large-scale datasets like ImageNet reveals a potential challenge. These datasets are characterized by a high density of semantically similar instances, which can lead {\mname} to enforce an excessive separation between them in the feature space. This over-separation risks corrupting the global semantic structure of the learned representations. Consequently, the size of the memory bank emerges as a critical hyperparameter that dictates the final performance.
{
    \small
    \bibliographystyle{ieeenat_fullname}
    \bibliography{main}
}

% WARNING: Do not forget to delete the supplementary pages from your submission 
\clearpage
\setcounter{page}{1}
\maketitlesupplementary

\section{Additional Experimental Details}
\label{App:Additional Experimental Details}

\begin{table}[t]
  \centering
  \small
  \begin{tabular}{lcccc}
    \toprule
    model & S/2 & B/2 & L/2 & XL/2 \\
    \midrule
    \multicolumn{5}{l}{\bfseries model configurations} \\
    params (M)           &  33  & 130  & 458  & 675   \\
    depth                &  12  &  12  &  24  &  28        \\
    hidden dim           & 384  & 768  &1024  &1152       \\
    patch size           &  2   &  2   &  2   &  2     \\
    heads                &  6   & 12   & 16   & 16  \\
    \midrule
    \multicolumn{5}{l}{\bfseries training configurations} \\
    epochs & 80 & 80 & 80 & 80 - 200 \\
    batch size           & \multicolumn{4}{c}{256}   \\
    optimizer            & \multicolumn{4}{c}{AdamW}  \\
    optimizer  $\beta_1$          & \multicolumn{4}{c}{0.9}  \\
    optimizer  $\beta_2$          & \multicolumn{4}{c}{0.95}  \\
    weight decay         & \multicolumn{4}{c}{0.0}     \\
    learning rate (lr)   & \multicolumn{4}{c}{$1\times10^{-4}$}     \\
    lr schedule          & \multicolumn{4}{c}{constant}      \\
    lr warmup            & \multicolumn{4}{c}{none}      \\
    \midrule
    \multicolumn{5}{l}{\bfseries ODE sampling} \\
    steps                  &      \multicolumn{4}{c}{250}   \\
    sampler              &     \multicolumn{4}{c}{Heun}  \\
    $t$ schedule        &     \multicolumn{4}{c}{linear}  \\
    last step size      & \multicolumn{4}{c}{N/A}  \\
    \midrule
    \multicolumn{5}{l}{\bfseries Dispersive Loss} \\
    regularization strength $\gamma$ &      \multicolumn{4}{c}{0.25}   \\
    temperature $\tau$ &      \multicolumn{4}{c}{0.5}   \\
    memory bank size $K$ &   65536   & 131072  & 32768 & 32768   \\
    \bottomrule
  \end{tabular}
  \vspace{.5em}
  \caption{\textbf{SiT Configuration on ImageNet.}
  }
  \label{tab:combined_configs}
\end{table}

Our ImageNet experiments are run on 8 A100 (80GB) GPUs. We strictly adhere to the settings and codebase of the SiT paper. Specifically, we utilize the AdamW optimizer with a learning rate of $1 \times 10^{-4}$, $(\beta_1, \beta_2) = (0.9, 0.95)$, and zero weight decay. The ODE sampling is performed using a Heun solver with 250 steps. Complete configuration details are available in Table \ref{tab:combined_configs}.

\section{Evaluation Detail}
\label{appen:eval}

We assess our method using several standard metrics:

\begin{itemize}[leftmargin=0.2in]
\item Fréchet Inception Distance (FID)~\citep{heusel2017gans}: This metric calculates the feature distance between the real and generated image distributions. It utilizes features extracted from the Inception-v3 network~\citep{szegedy2016rethinking}, assuming both feature sets follow a multivariate Gaussian distribution.
\item Spatial FID (sFID)~\citep{nash2021generating}: A variant of FID, sFID computes the distance using intermediate spatial features from the Inception-v3 network, specifically to evaluate the spatial characteristics of the generated images.
\item Precision~\citep{kynkaanniemi2019improved}: Adapted from their classic definitions, precision measures the fraction of generated images considered realistic, while recall measures the fraction of the true data manifold covered by the generated distribution.
\item Inception Score (IS)~\citep{salimans2016improved}: Also leveraging the Inception-v3 network, IS evaluates quality and diversity by measuring the KL-divergence between the conditional label distribution (derived from softmax-normalized logits) and the marginal label distribution.
\end{itemize}

\section{More qualitative results}

\begin{figure*}
  \centering
  \includegraphics[width=\textwidth]{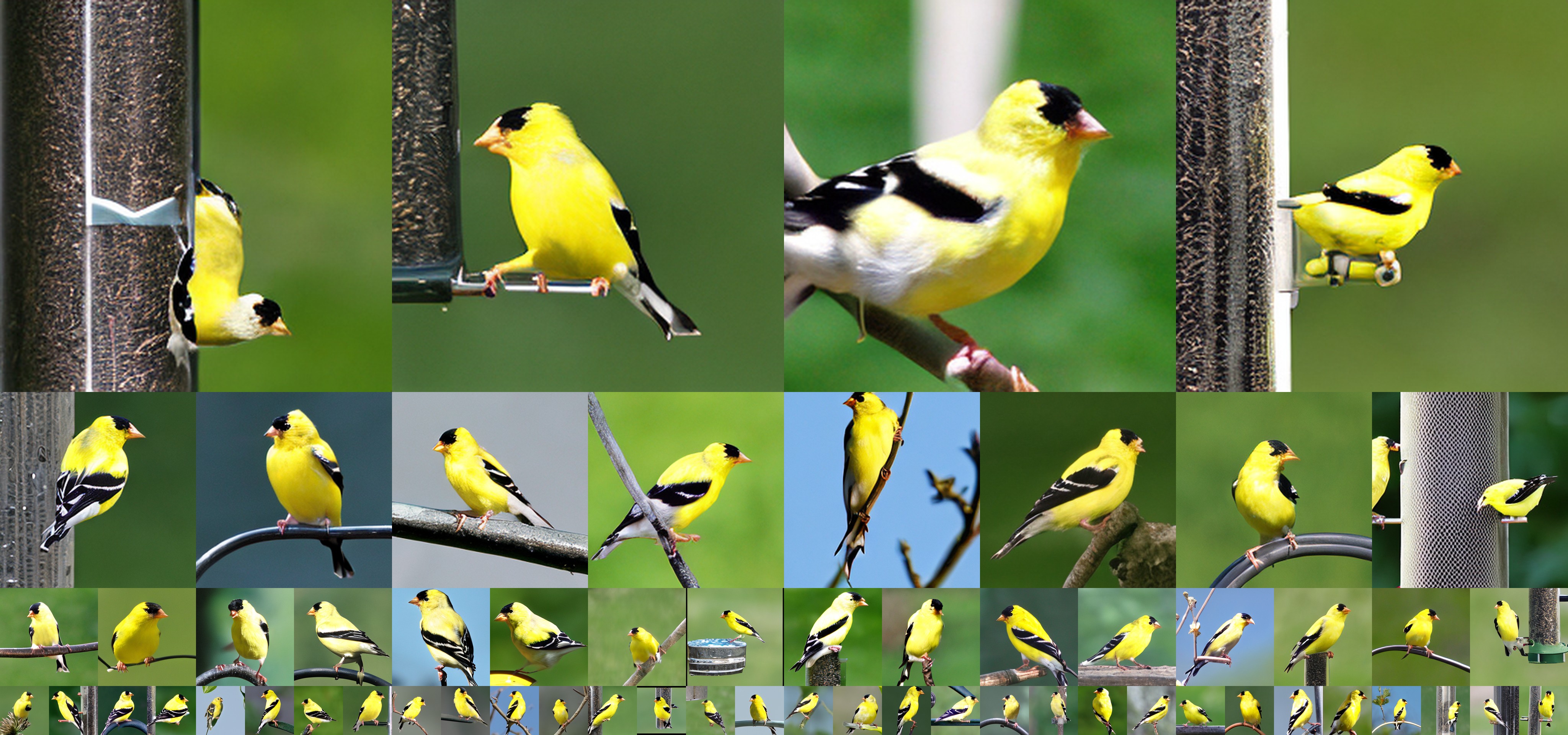}
  \caption{Uncurated generation results of SiT-XL/2+{\mname}. We use classifier-free guidance with $w=4.0$. Class label=“goldfinch”(11).}
\end{figure*}

\begin{figure*}
  \centering
  \includegraphics[width=\textwidth]{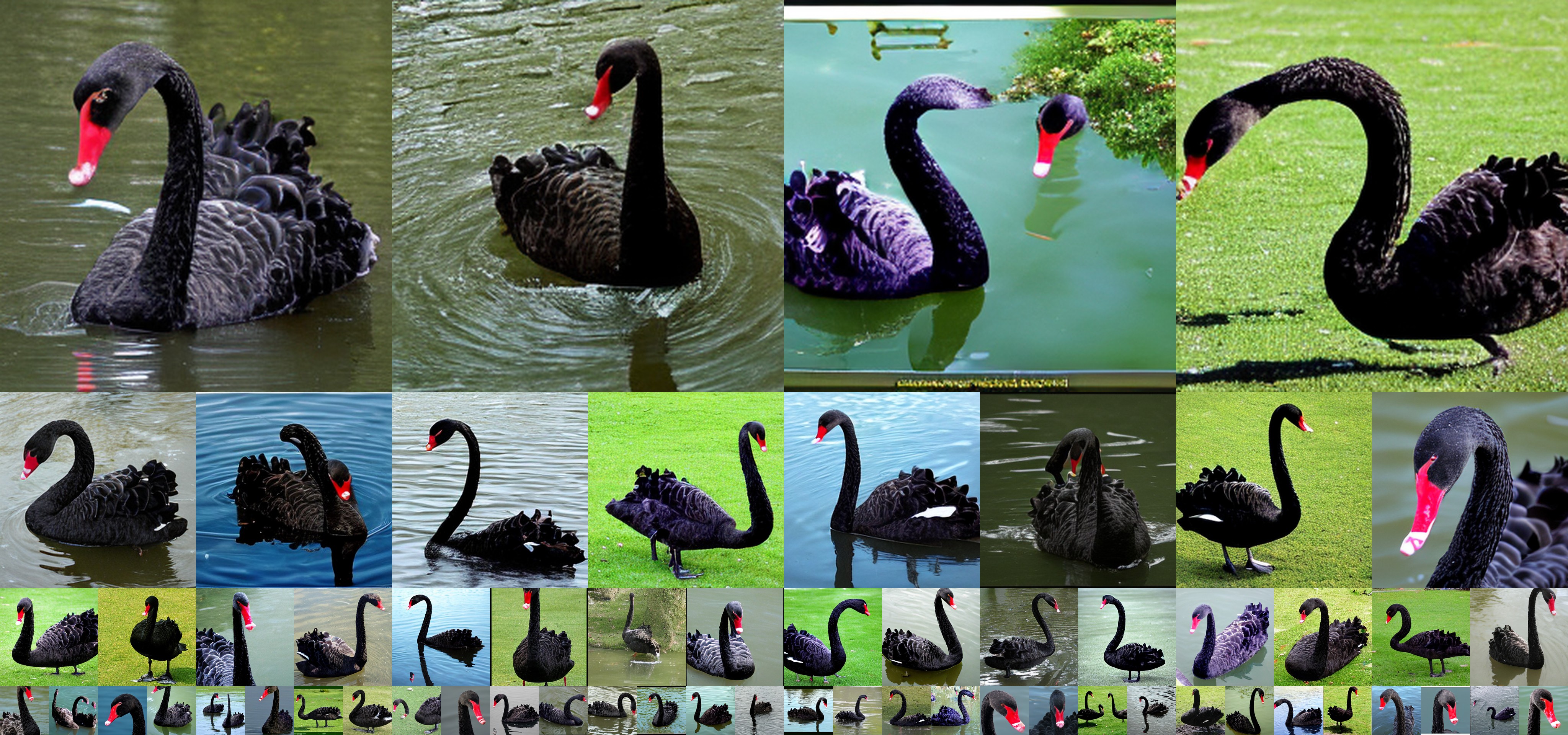}
  \caption{Uncurated generation results of SiT-XL/2+{\mname}. We use classifier-free guidance with $w=4.0$. Class label=“black swan”(100).}
\end{figure*}

\begin{figure*}
  \centering
  \includegraphics[width=\textwidth]{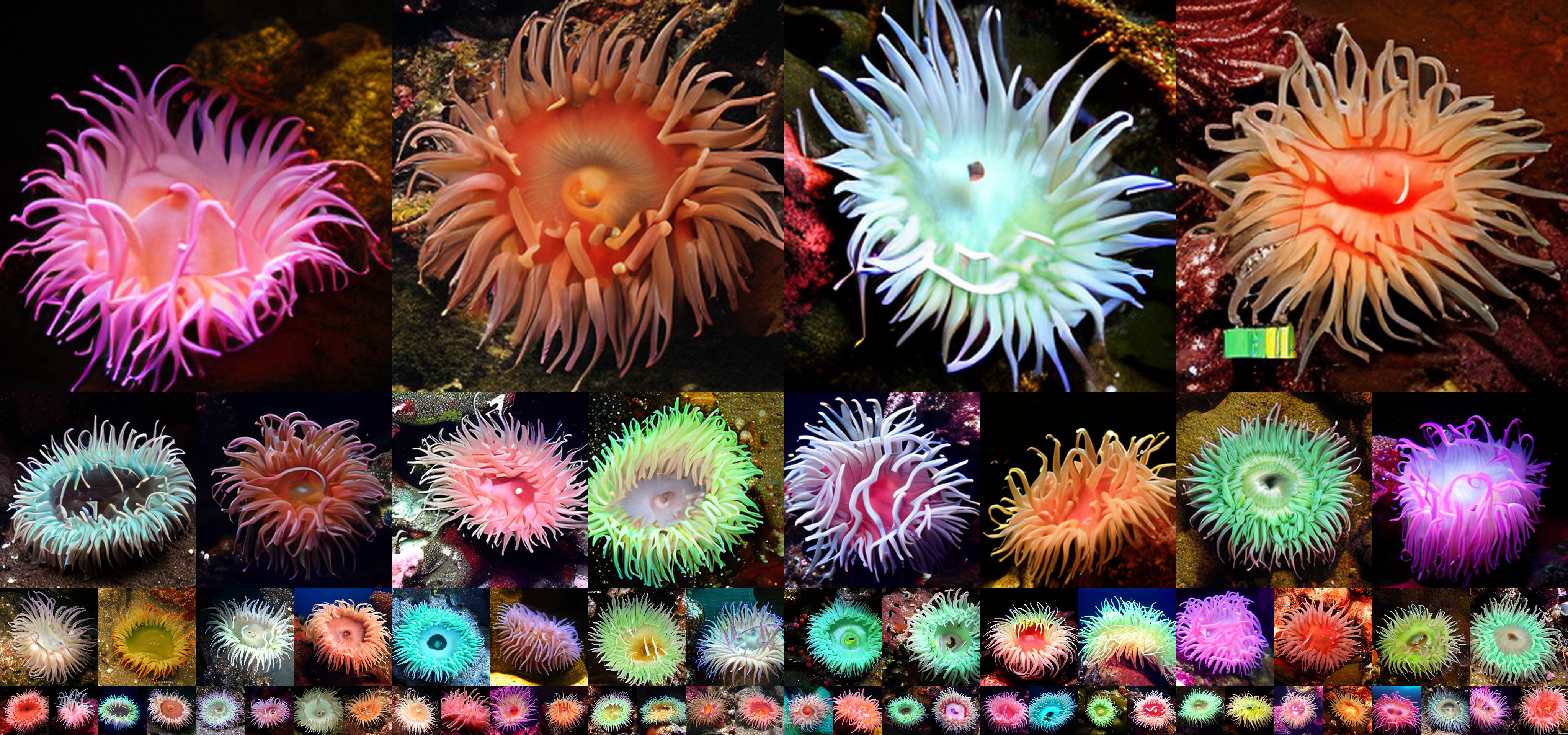}
  \caption{Uncurated generation results of SiT-XL/2+{\mname}. We use classifier-free guidance with $w=4.0$. Class label=“sea anemone”(108).}
\end{figure*}

\begin{figure*}
  \centering
  \includegraphics[width=\textwidth]{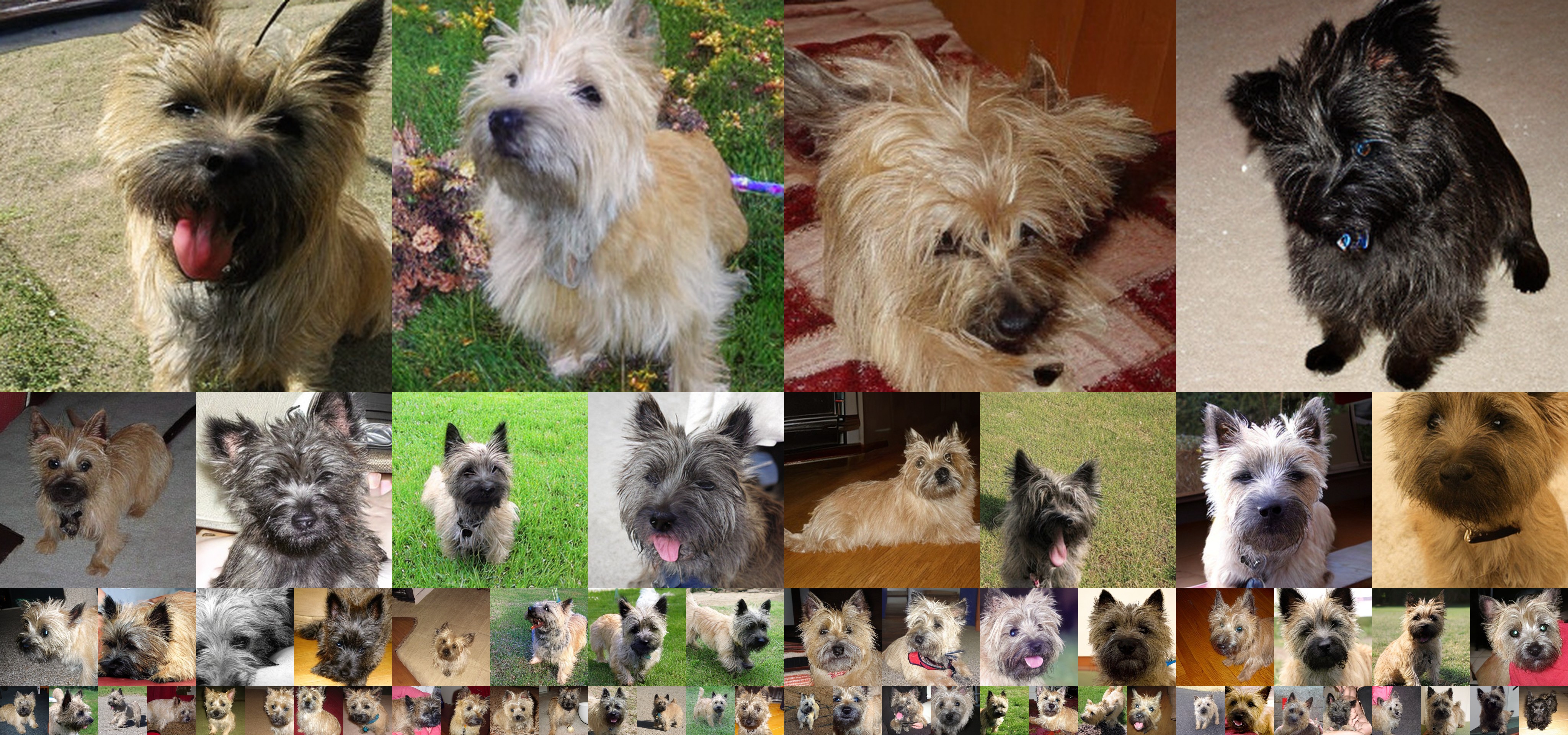}
  \caption{Uncurated generation results of SiT-XL/2+{\mname}. We use classifier-free guidance with $w=4.0$. Class label=“cairn”(192).}
\end{figure*}

\begin{figure*}
  \centering
  \includegraphics[width=\textwidth]{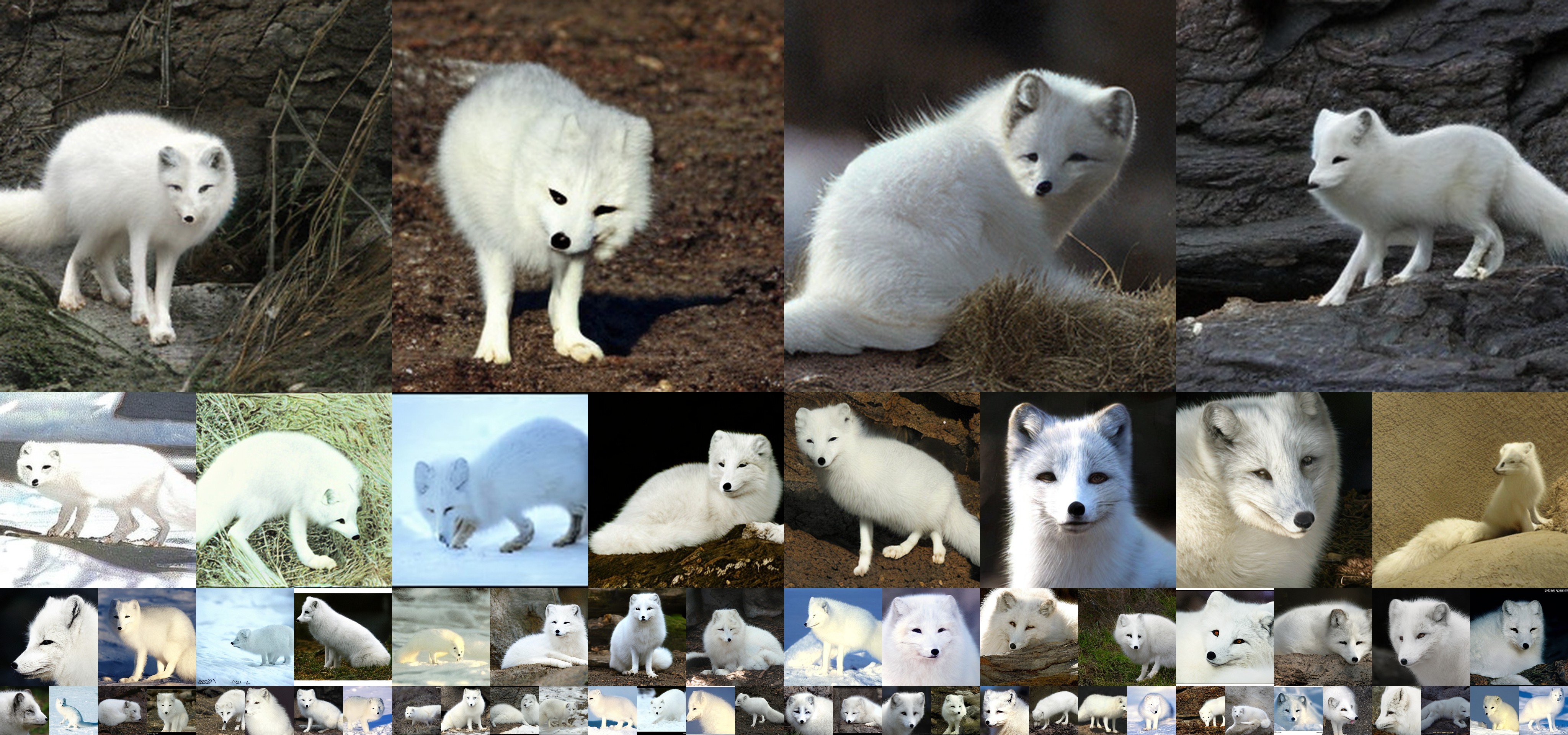}
  \caption{Uncurated generation results of SiT-XL/2+{\mname}. We use classifier-free guidance with $w=4.0$. Class label=“arctic fox”(279).}
\end{figure*}

\begin{figure*}
  \centering
  \includegraphics[width=\textwidth]{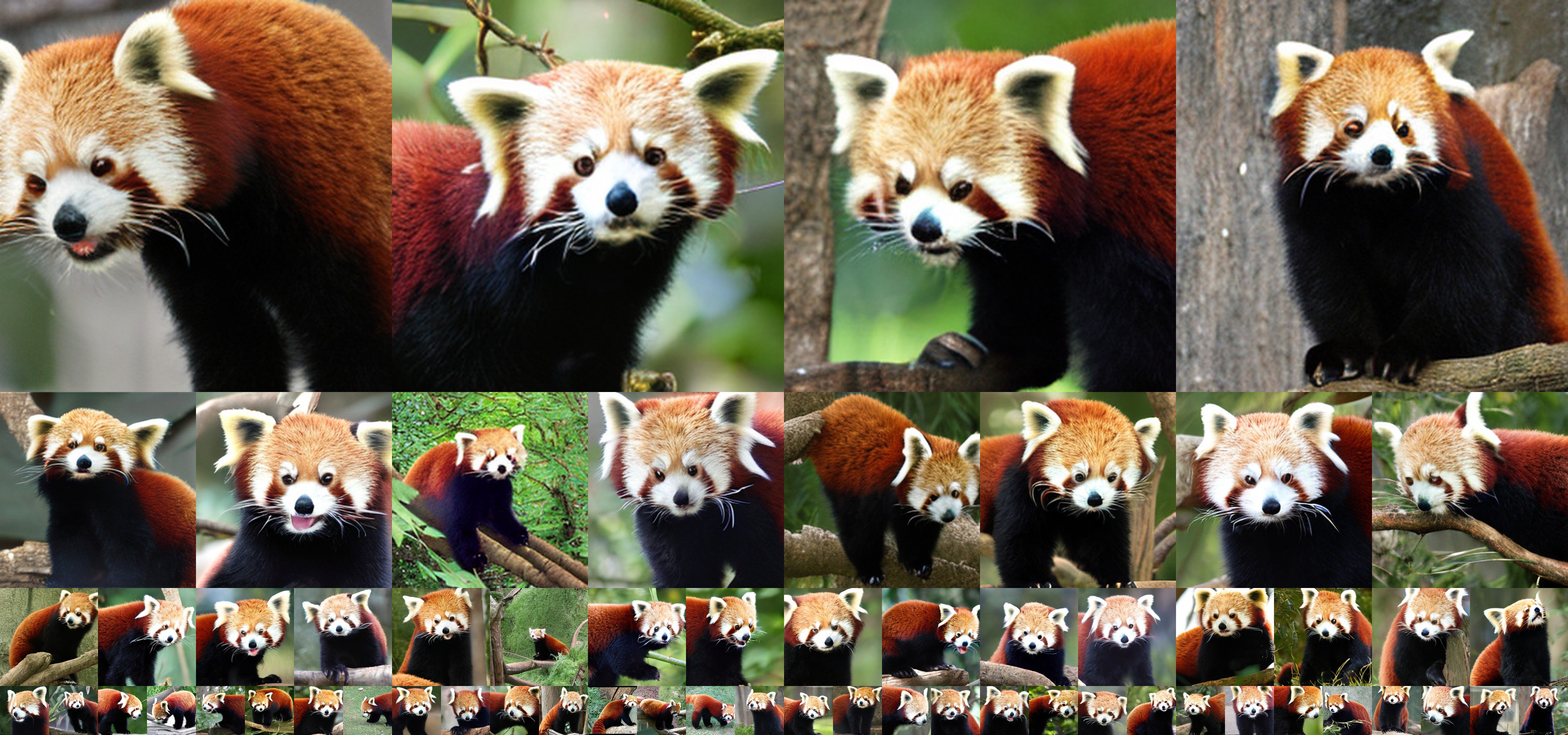}
  \caption{Uncurated generation results of SiT-XL/2+{\mname}. We use classifier-free guidance with $w=4.0$. Class label=“lesser panda”(387).}
\end{figure*}

\begin{figure*}
  \centering
  \includegraphics[width=\textwidth]{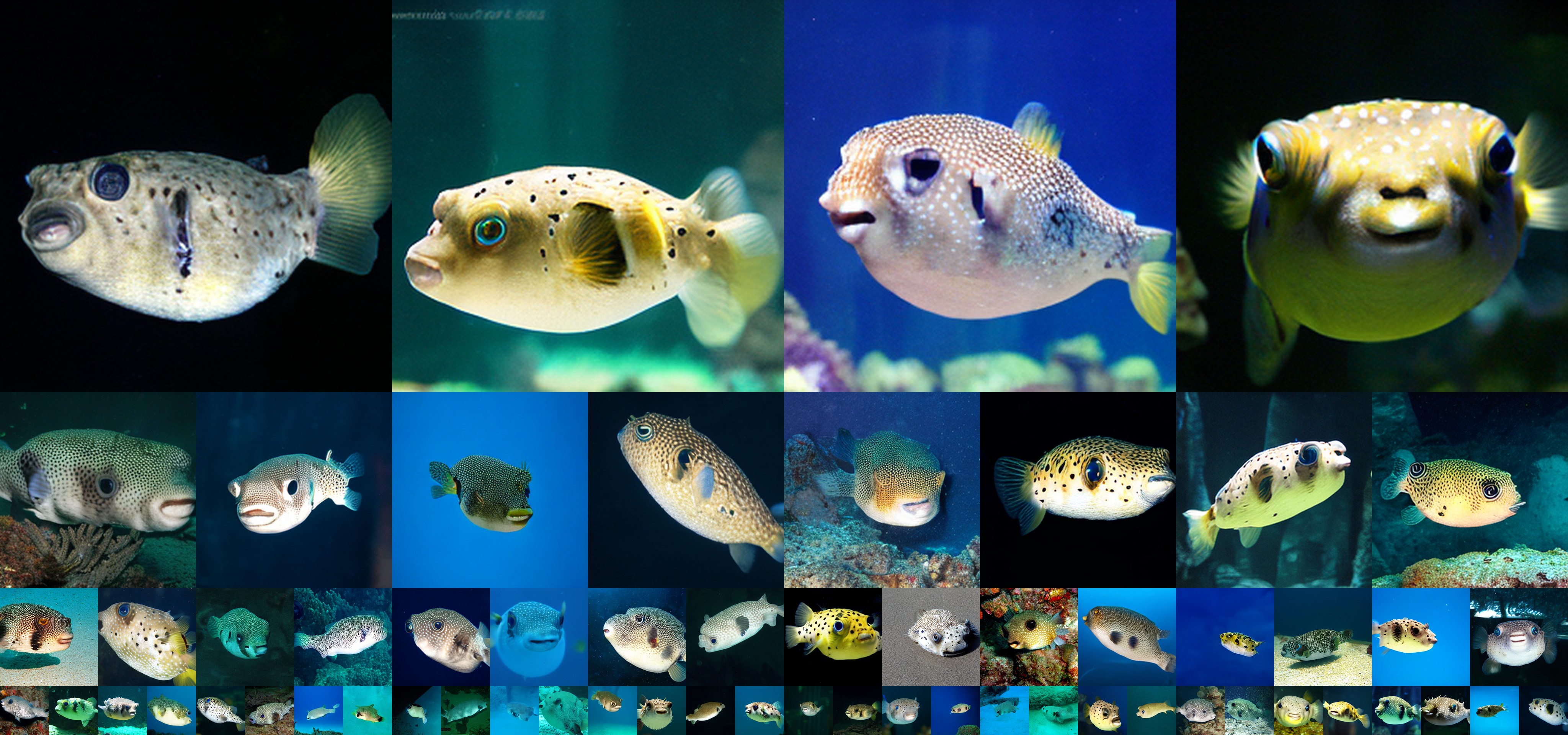}
  \caption{Uncurated generation results of SiT-XL/2+{\mname}. We use classifier-free guidance with $w=4.0$. Class label=“puffer”(397).}
\end{figure*}

\begin{figure*}
  \centering
  \includegraphics[width=\textwidth]{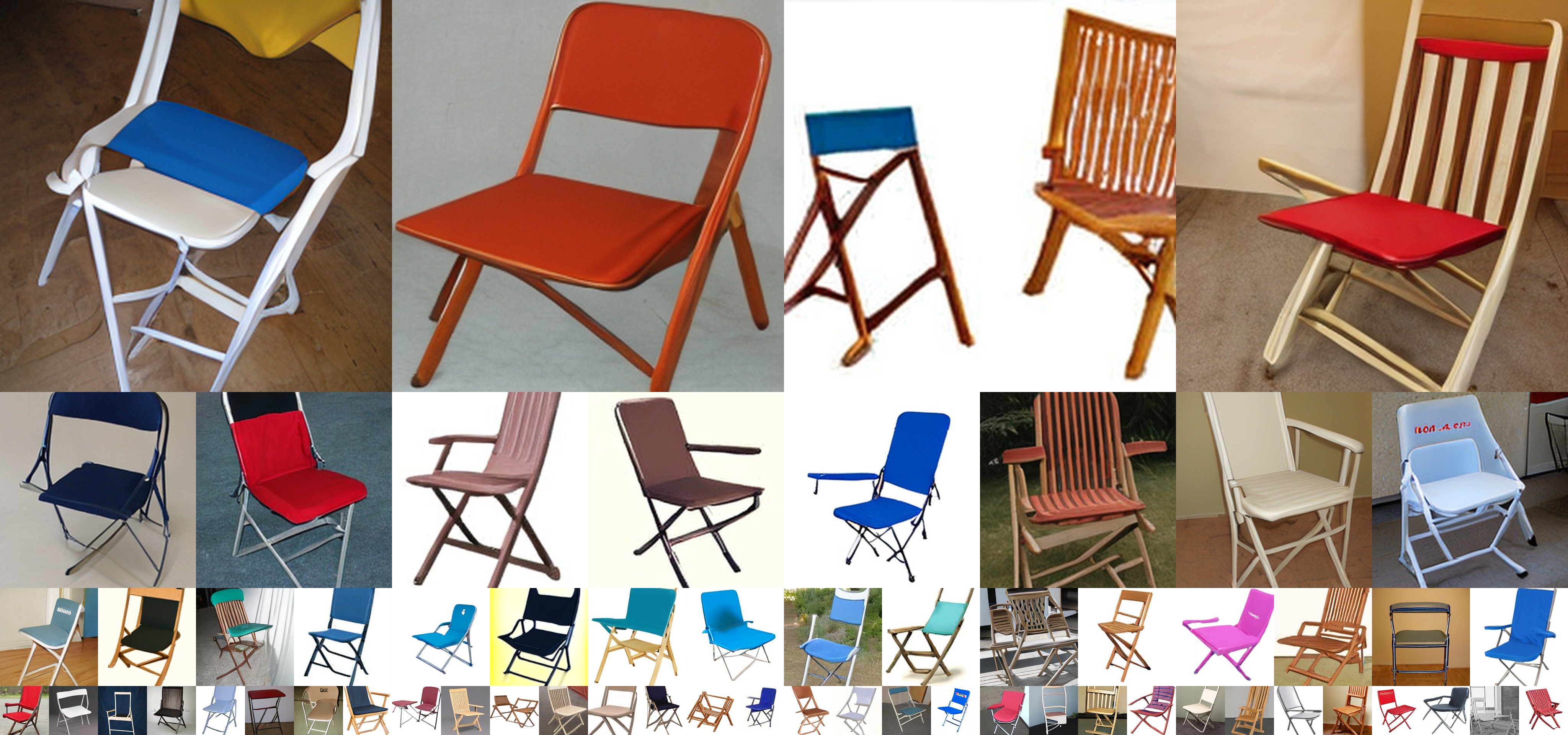}
  \caption{Uncurated generation results of SiT-XL/2+{\mname}. We use classifier-free guidance with $w=4.0$. Class label=“folding chair”(559).}
\end{figure*}

\begin{figure*}
  \centering
  \includegraphics[width=\textwidth]{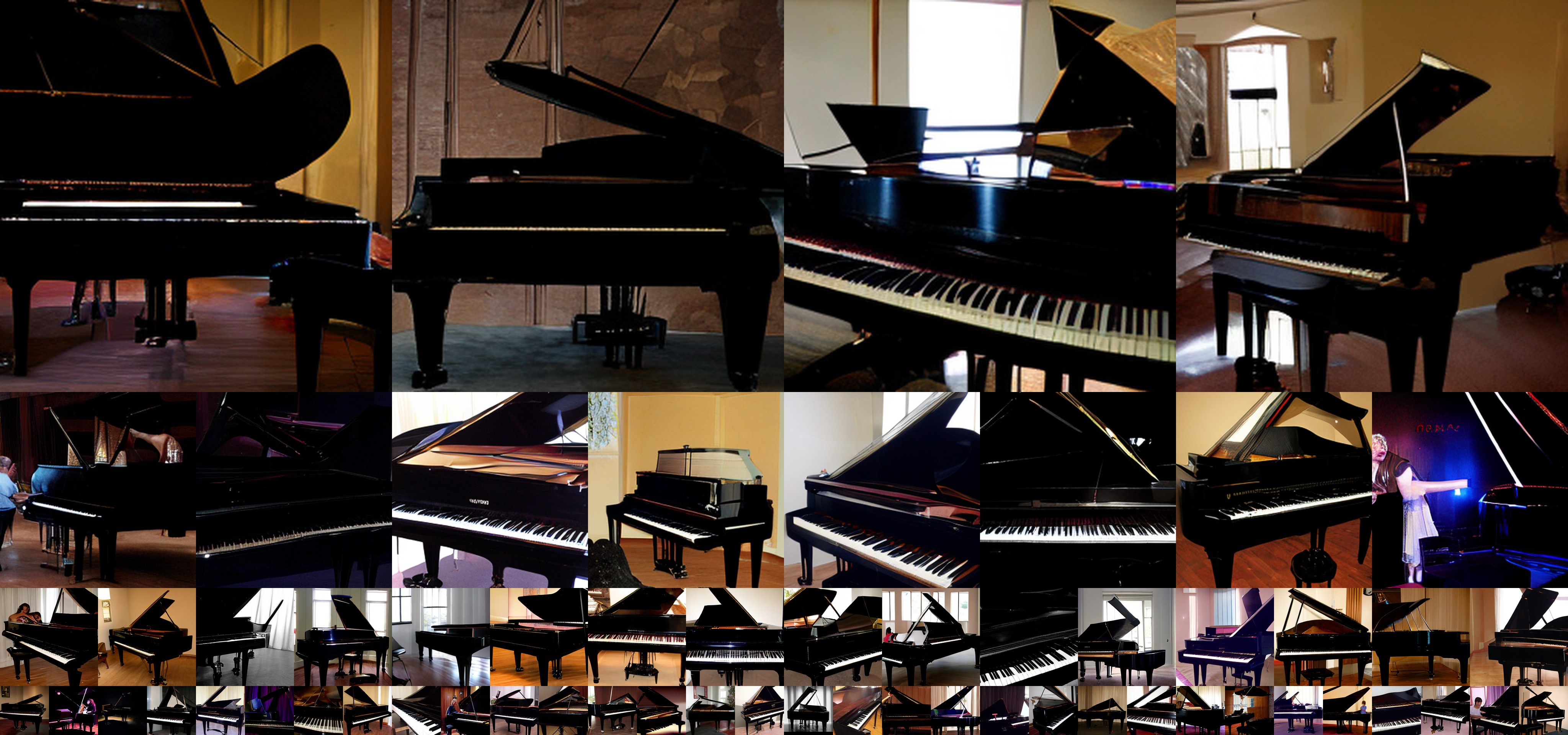}
  \caption{Uncurated generation results of SiT-XL/2+{\mname}. We use classifier-free guidance with $w=4.0$. Class label=“grand piano”(579).}
\end{figure*}

\begin{figure*}
  \centering
  \includegraphics[width=\textwidth]{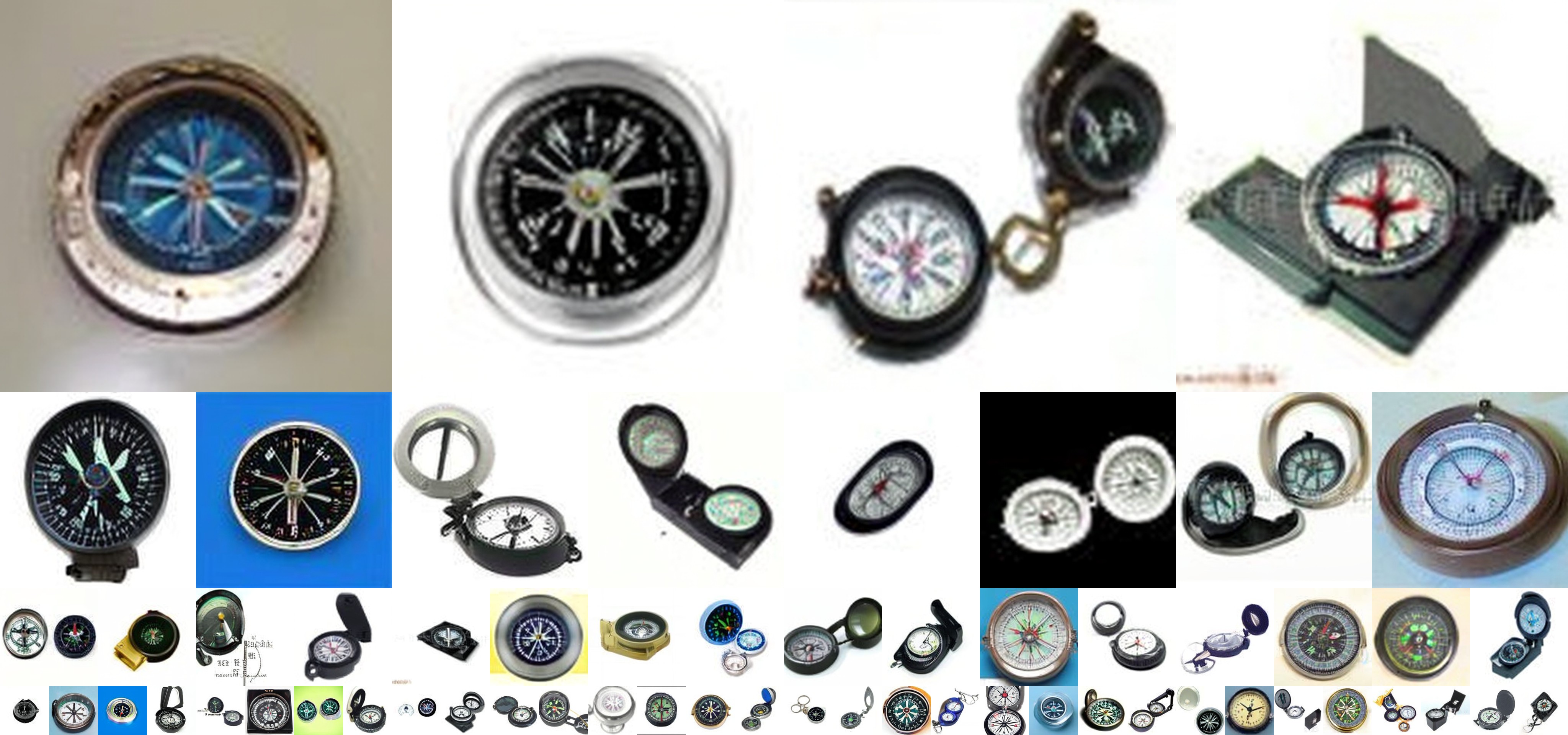}
  \caption{Uncurated generation results of SiT-XL/2+{\mname}. We use classifier-free guidance with $w=4.0$. Class label=“magnetic compass”(635).}
\end{figure*}

\begin{figure*}
  \centering
  \includegraphics[width=\textwidth]{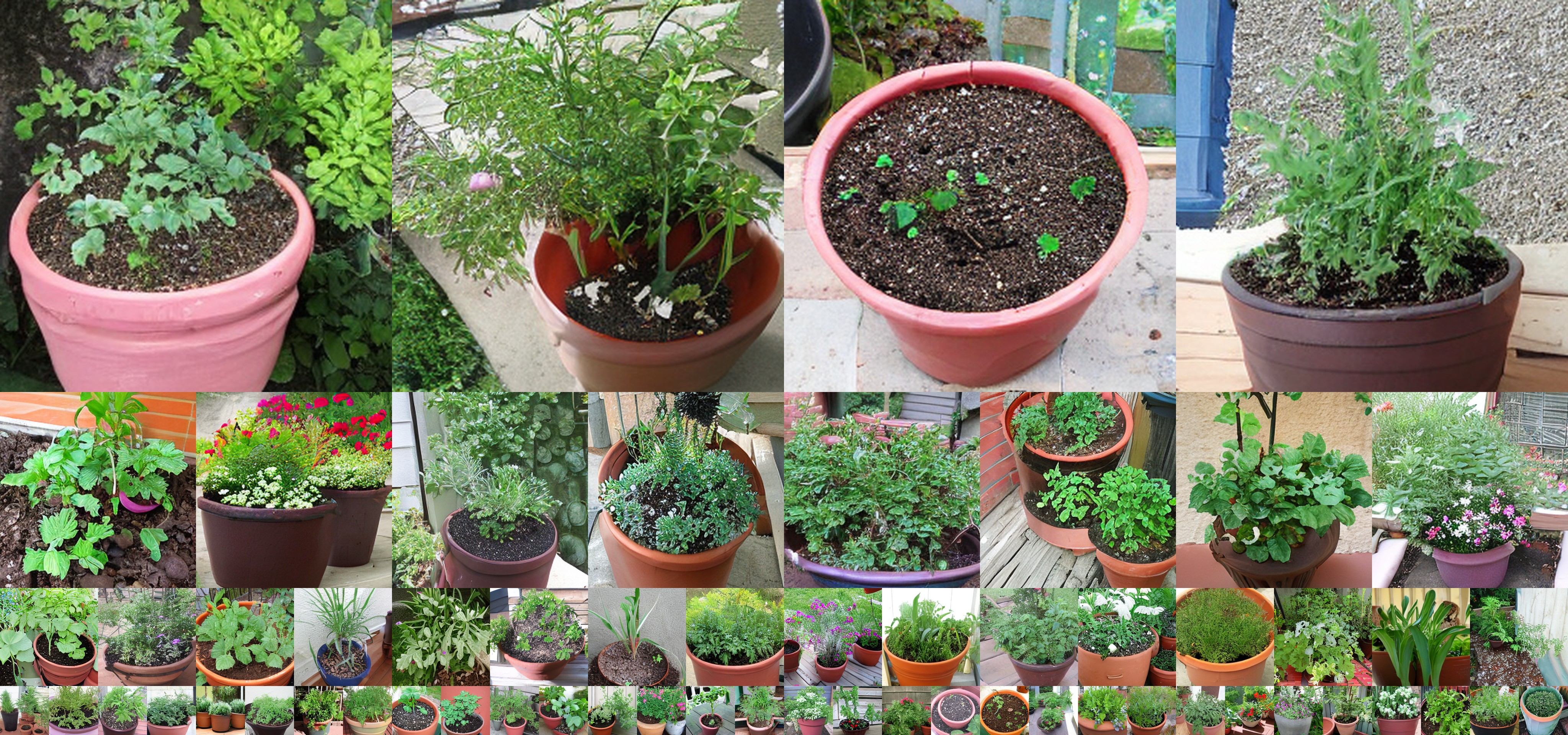}
  \caption{Uncurated generation results of SiT-XL/2+{\mname}. We use classifier-free guidance with $w=4.0$. Class label=“pot”(738).}
\end{figure*}

\begin{figure*}
  \centering
  \includegraphics[width=\textwidth]{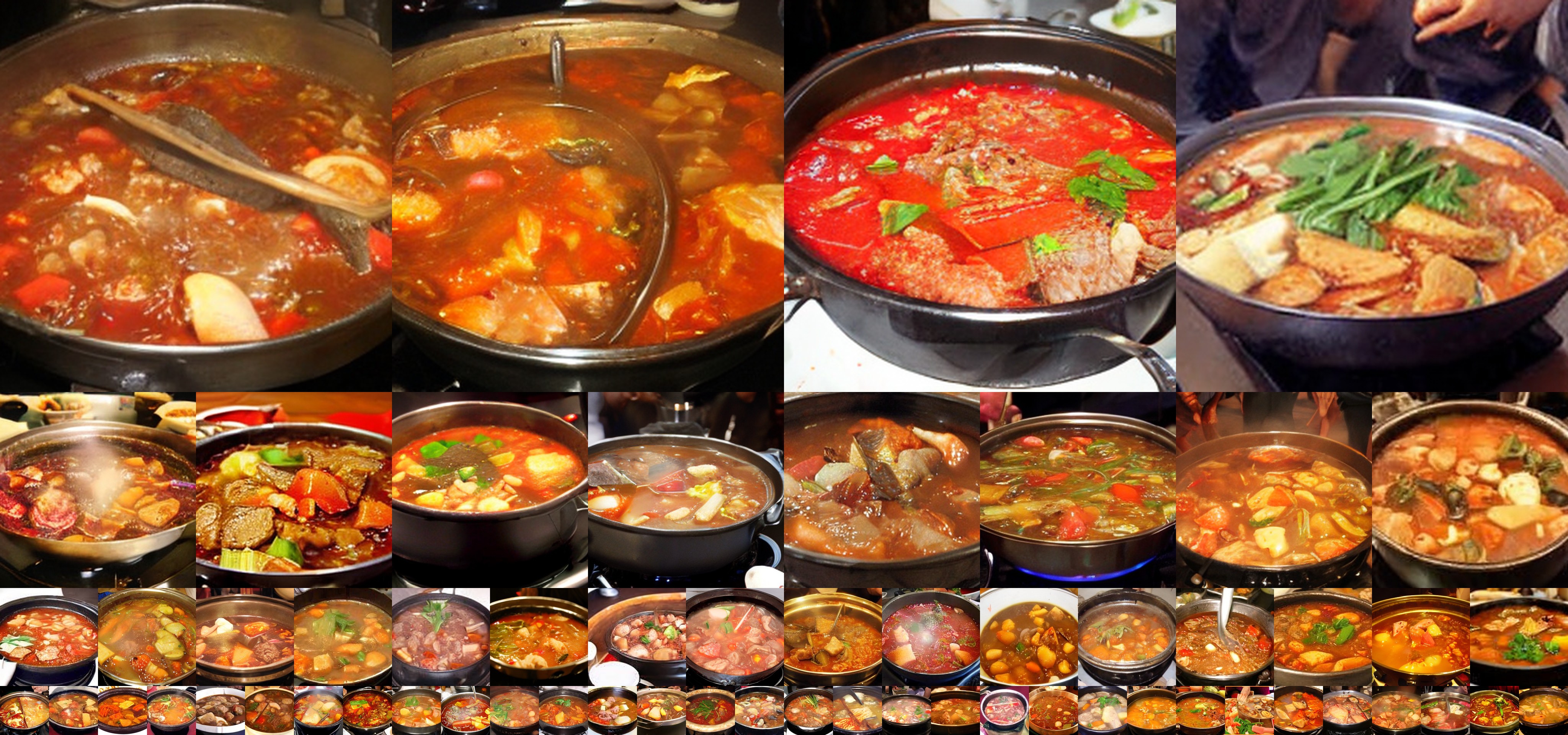}
  \caption{Uncurated generation results of SiT-XL/2+{\mname}. We use classifier-free guidance with $w=4.0$. Class label=“hot pot”(926).}
\end{figure*}

\begin{figure*}
  \centering
  \includegraphics[width=\textwidth]{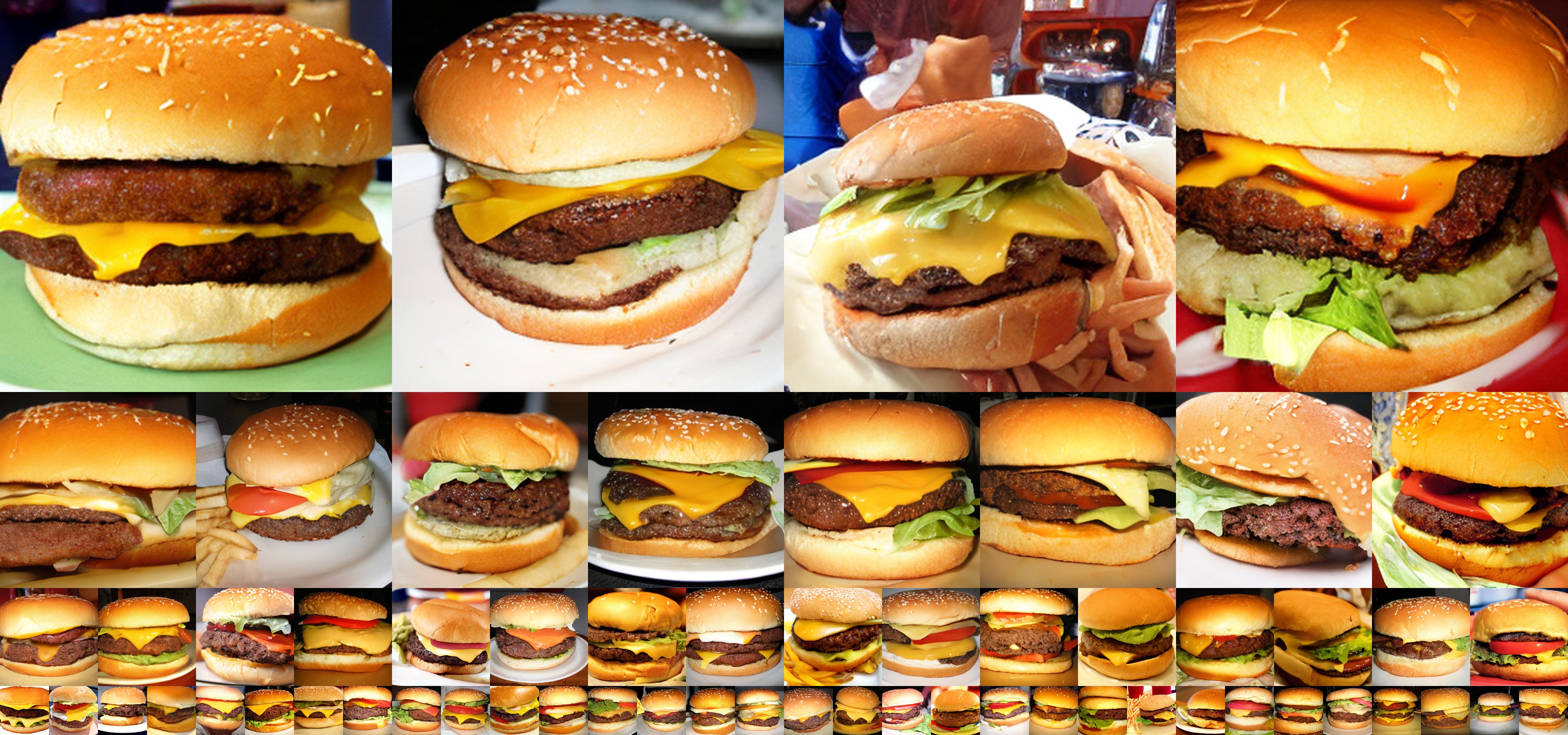}
  \caption{Uncurated generation results of SiT-XL/2+{\mname}. We use classifier-free guidance with $w=4.0$. Class label=“cheeseburger”(933).}
\end{figure*}

\begin{figure*}
  \centering
  \includegraphics[width=\textwidth]{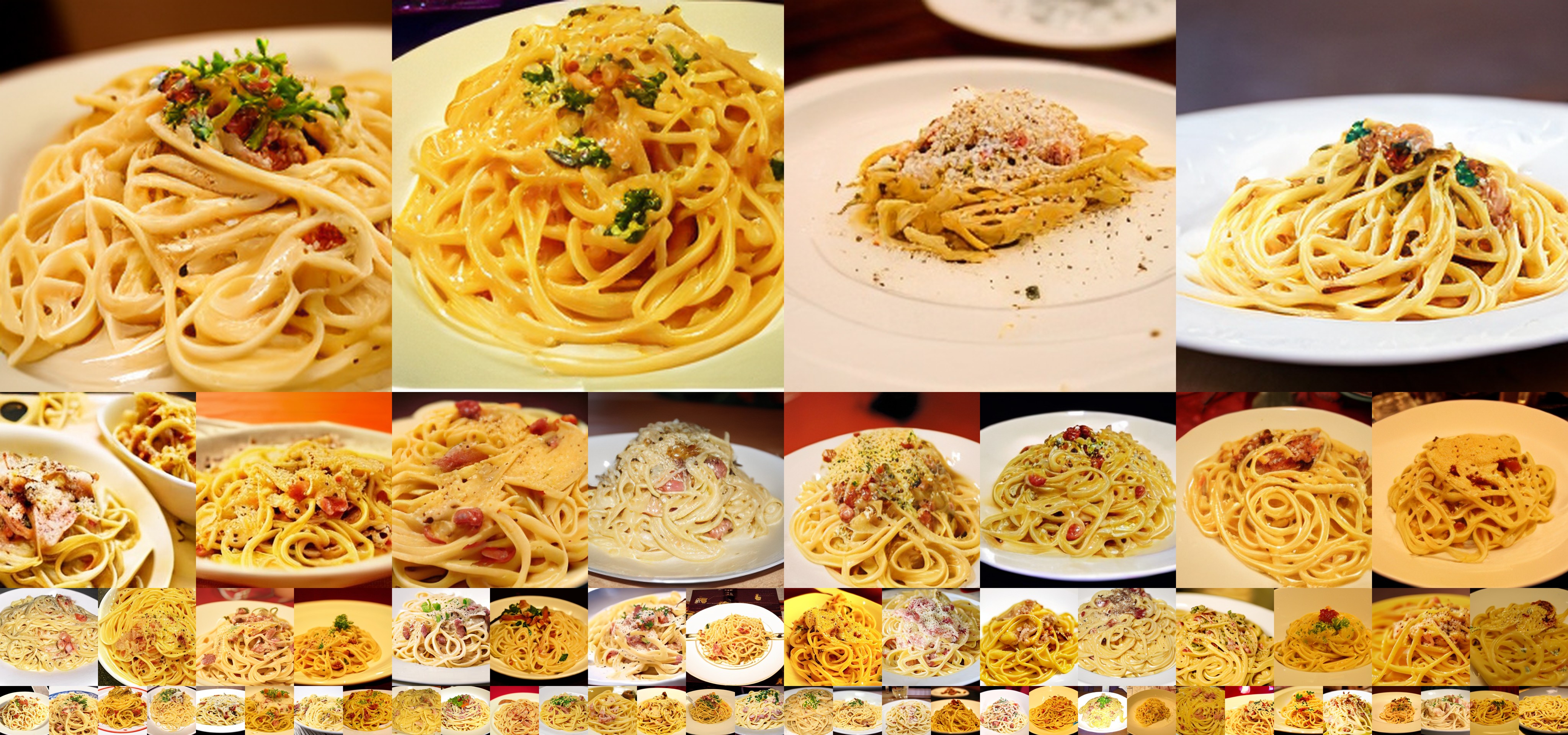}
  \caption{Uncurated generation results of SiT-XL/2+{\mname}. We use classifier-free guidance with $w=4.0$. Class label=“carbonara”(959).}
\end{figure*}

\end{document}